\documentclass[lettersize,journal]{IEEEtran}
\usepackage{amsmath,amsfonts}
\usepackage{algorithmic}
\usepackage{algorithm}
\usepackage{array}
\usepackage[caption=false,font=normalsize,labelfont=sf,textfont=sf]{subfig}
\usepackage{textcomp}
\usepackage{stfloats}
\usepackage{url}
\usepackage{verbatim}
\usepackage{graphicx}
\usepackage{colortbl}
\usepackage{cite}
\usepackage[table]{xcolor}
\usepackage{booktabs}
\usepackage{tabularx}
\usepackage{multirow}
\usepackage{hyperref}
\newcommand{\textpink}[1]{\textcolor[rgb]{0.85,0.2,0.7}{#1}}
\newcommand{\textgreen}[1]{\textcolor{green!60!black}{#1}}
\newcommand{\textred}[1]{\textcolor{red!70!black}{#1}}

\usepackage[most]{tcolorbox}
\definecolor{titleblue}{RGB}{0,51,102}
\definecolor{boxblue}{RGB}{230,240,255}

\newtcolorbox{hlgreenbox}{
  colback=green!20,
  colframe=green!20,
  boxrule=0pt,
  arc=2pt,
  left=4pt,right=4pt,top=2pt,bottom=2pt
}

\newtcolorbox{hlredbox}{
  colback=red!20,
  colframe=red!20,
  boxrule=0pt,
  arc=2pt,
  left=4pt,right=4pt,top=2pt,bottom=2pt
}

\newtcolorbox{promptbox}[1]{
  coltitle=white,         
  boxrule=1pt,            
  arc=4pt,                
  left=8pt, right=8pt,    
  top=8pt, bottom=8pt,
  breakable,              
  enhanced,               
  fonttitle=\normalfont,  
  title={#1}              
}

\begin{document}

\title{Think Thrice Before You Speak: \\ Dual knowledge-enhanced Theory-of-Mind Reasoning for Persuasive Agents}


\author{
Minghui Ma, 
Bin Guo*, \IEEEmembership{Senior Member, IEEE}, 
Runze Yang, 
Mengqi Chen, 
Yan Liu, 
Jingqi Liu, 
Yahan Pei, 
Xuehao Ma, 
Qiuyun Zhang, 
Zhiwen Yu, \IEEEmembership{Senior Member, IEEE}%
\thanks{This work was partially supported by the National Natural Science Foundation of China (Nos. U25B2042, 62532009, 62502392).}%
\thanks{Minghui Ma, Bin Guo*, Runze Yang, Mengqi Chen, Yan Liu, Jingqi Liu, Yahan Pei, and Xuehao Ma are with Northwestern Polytechnical University, Xi'an, China (e-mail: \{2021300120, yangrunze, chenmengqi, 1309893591, peiyahan, xhma\}@mail.nwpu.edu.cn; guob@nwpu.edu.cn; liu.yan@nwpu.edu.cn).}%
\thanks{Qiuyun Zhang is with Peking University, Beijing, China (e-mail: qiuyunzhang@pku.edu.cn). %
Zhiwen Yu is with Harbin Engineering University, Harbin, China, and Northwestern Polytechnical University, Xi'an, China (e-mail: zhiwenyu@nwpu.edu.cn).}%
\thanks{*Corresponding author: Bin Guo (e-mail: guob@nwpu.edu.cn).}%
}

\markboth{Journal of \LaTeX\ Class Files,~Vol.~14, No.~8, August~2021}%
{Shell \MakeLowercase{\textit{et al.}}: A Sample Article Using IEEEtran.cls for IEEE Journals}


\maketitle

\begin{abstract}
Persuasive dialogue requires reasoning about others’ latent mental states, a capability known as \emph{Theory of Mind} (ToM). However, due to reliance on simple prompting strategies and insufficient ToM knowledge, existing LLMs often fail to capture the intrinsic dependencies among mental states, leading to fragmented representations and unstable reasoning. To address these challenges, we introduce the \textbf{ToM-based Persuasive Dialogue (ToM-PD)} task, grounded in the \textit{Belief--Desire--Intention} (BDI) framework, which explicitly models the sequential dependencies among mental states in multi-turn dialogues. To facilitate research on this task, we construct a large-scale annotated dataset, \textbf{ToM-based Broad Persuasive Dialogues (ToM-BPD)}, capturing fine-grained mental states and corresponding persuasive strategies. We further propose \textbf{Think Thrice Before You Speak (TTBYS)}, a knowledge-enhanced stepwise reasoning framework that leverages both explicit and implicit prior experiences to improve LLMs’ inference of desires, beliefs, and strategies. Experimental results demonstrate that Qwen3-8B equipped with TTBYS outperforms GPT-5 by 1.20\%, 22.80\%, and 16.97\% in predicting desires, beliefs, and persuasive strategies, respectively. Case studies further show that our approach enhances interpretability and consistency in reasoning. 

\end{abstract}

\begin{figure*}[t]
  \centering
  \includegraphics[width=0.96\linewidth]{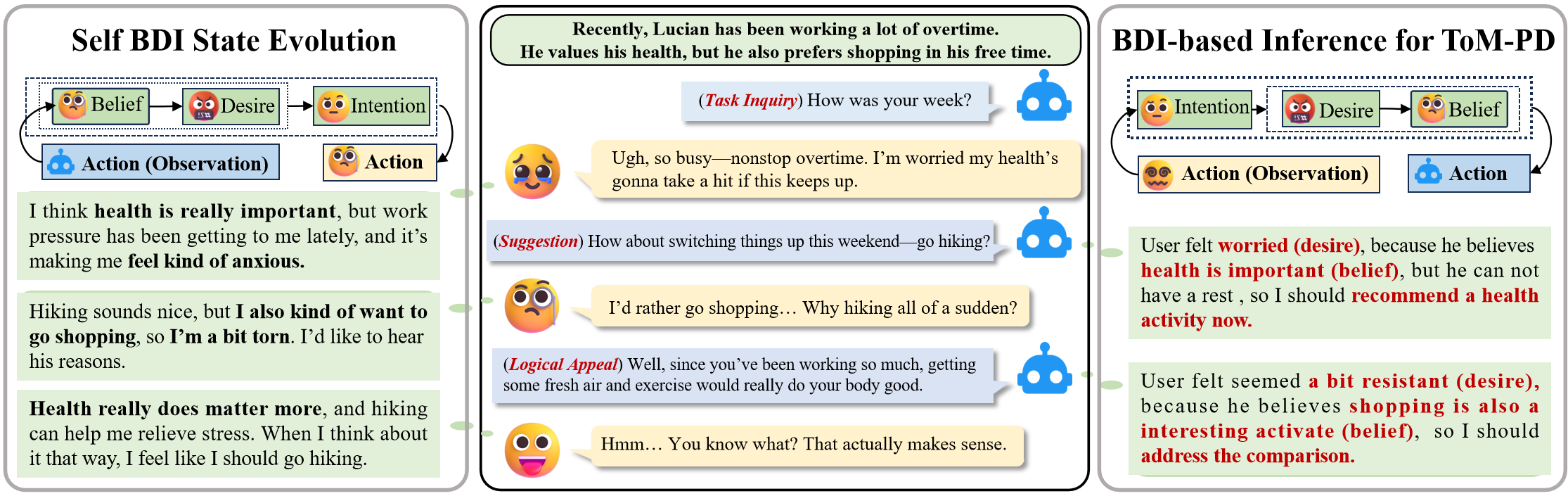}
  \caption{Illustration of self BDI state evolution and BDI-based inference for ToM-driven persuasive dialogue (ToM-PD). The left panel shows the internal reasoning process, where Lucian generates actions through the evolution of its belief, desire, and intention states based on self-perception and experience. The middle panel presents a multi-turn persuasive dialogue scenario between the agent and the user. The right panel depicts the ToM-PD process, where the agent observes user actions (utterances), infers the user’s latent BDI states, and dynamically selects appropriate persuasive strategies to guide subsequent actions.}
  \label{fig:fig1}
\end{figure*}


\section{Introduction}
\IEEEPARstart{P}{ersuasion} is a fundamental component of human social interaction \cite{petty1986elaboration,petty2012communication}. With the rapid advancement of LLMs, LLM-based persuasive dialogue systems have gained increasing influence across diverse domains, such as emotional support \cite{liu2021towards,zhang2024escot}, charitable donation \cite{mishra2022pepds,song2024would}, and negotiation \cite{kwon2024llms}. These systems not only need to generate fluent and coherent responses but also accurately infer and strategically influence users’ latent mental states\cite{han2025tomaptrainingopponentawarellm}.

Theory of Mind, the ability to infer others’ beliefs, desires, and intentions, is widely recognized as a core cognitive mechanism underlying social reasoning and strategic interaction \cite{premack1978does,baron1985does}. Motivated by this insight, recent studies have explored the integration of ToM modeling into persuasive dialogue systems. For instance, prior work decomposes persuasion tasks into multiple collaborating agents, with dedicated components for identifying users’ emotional states \cite{cheng2024cooper}. Other approaches simulate potential user rebuttals and attitudes to improve persuasive effectiveness \cite{han2025tomaptrainingopponentawarellm}. In addition, chain-of-thought reasoning has been leveraged to model users’ mental states and enhance the empathy of system responses \cite{zhang2024escot, chang2024injectingsalespersonsdialoguestrategies}. Although effective on benchmark tasks, these approaches often treat different mental states as independent variables, neglecting their intrinsic dependencies, which can lead to inconsistent or unstable inferences in complex dialogues.

Several benchmarks have also evaluated LLMs’ ToM reasoning capabilities in persuasive contexts. NegotiationToM \cite{chan2024negotiationtom} focuses on negotiation scenarios, ToMATO \cite{shinoda2025tomato} targets donation requests, and BigToM \cite{gandhi2023understanding} evaluates action prediction in narrative contexts. PersuasiveToM \cite{yu2025persuasivetom} further examines LLMs’ ability to reason about and predict persuasive strategies based on users’ mental states. However, empirical results consistently indicate that LLMs lack sufficient prior knowledge for reliable mental-state inference, limiting the stability and robustness of their reasoning. 

In summary, applying ToM to persuasive dialogue systems and improving their performance still faces two major challenges: (1) how to systematically model and infer users’ ToM states and their intrinsic dependencies in persuasive dialogues, and (2) how to equip LLMs with sufficient knowledge to enhance their ToM reasoning capabilities.

As shown in Figure~\ref{fig:fig1}, the user’s internal BDI states and corresponding actions follow a forward evolutionary process, in which mental states evolve sequentially from beliefs to desires, intentions, and ultimately actions \cite{georgeff1998belief}. In contrast, we model persuasive interaction as an inference problem centered on first-order ToM. The agent can only observe the persuadee’s external utterances and must iteratively infer their intentions, further reasoning about the underlying desires and beliefs, in order to dynamically adjust persuasive strategies. Motivated by this observation and grounded in the BDI framework \cite{georgeff1998belief}, we formally define the \textbf{ToM-based Persuasive Dialogue (ToM-PD)} task, and formulate its solution as a stepwise backward inference process over latent mental states.

Humans often rely on prior experiences in analogous scenarios to guide reasoning when confronting ToM problems \cite{perner2007episodic}. In ToM-PD, this manifests as a stepwise, deliberate reasoning process, akin to the Chinese proverb, \textit{``Think Thrice Before You Act.”}. Drawing inspiration from this, we propose a knowledge-enhanced stepwise reasoning framework, \textbf{Think Thrice Before You Speak (TTBYS)}, which integrates both explicit and implicit knowledge to guide LLMs through sequential reasoning over desires, beliefs, and persuasive strategies, thereby improving robustness and consistency in ToM reasoning. To evaluate our approach, we further construct a multi-domain dataset, \textbf{ToM-based Broad Persuasive Dialogues (ToM-BPD)}, annotated with fine-grained ToM states of the persuadee and corresponding persuasive strategies of the persuader for each dialogue turn, with rigorous quality control to ensure annotation reliability.

In summary, our contributions are threefold: 

(1) We introduce the \textbf{ToM-PD} task, explicitly modeling dependencies among different mental states to enhance the interpretability.

(2) We propose \textbf{TTBYS}, a knowledge-enhanced stepwise reasoning framework to enhance ToM inference for LLM in persuasive dialogues.

(3) We construct a ToM-oriented persuasive dialogue dataset, \textbf{ToM-BPD}, and validate the effectiveness and interpretability of our approach through extensive experiments and case studies.

The rest of the paper is organized as follows. Section~\ref{sec:related} reviews related work on LLM-based persuasive dialogue agents and Theory of Mind in AI. Section~\ref{sec:tpmpd} formalizes the ToM-PD task and introduces our approach for sequential mental state inference in persuasive interactions. Section~\ref{sec:method} presents the \textbf{TTBYS} framework, detailing experience representation, retrieval mechanisms, and multi-step reasoning for desire, belief, and strategy inference. Section~\ref{sec:exp} presents experimental results and analysis. Section~\ref{sec:conclusion} draws conclusions, and Section~\ref{sec:ethics} provides an ethics statement.


\section{Related Work}\label{sec:related}
\subsection{LLM-based Persuasive Dialogue Agents}
Persuasive dialogue aims to influence individuals’ beliefs, attitudes, or behaviors through targeted communication strategies \cite{shi2020effects}. Early studies mainly focused on domain-specific scenarios such as emotional support \cite{liu2021towards}, policy persuasion \cite{chen2019multi}, and charitable fundraising \cite{wang2019persuasion}. With the rise of large language models, recent work has extended persuasive dialogue to more diverse and complex settings, including multi-turn recommendation \cite{kim2025towards}, adversarial prompting \cite{zeng2024johnny}, and misleading or manipulative behaviors \cite{xu2024earth, furumai2024zero}.

From a technical perspective, existing approaches can be broadly categorized into three lines. 
\textit{Strategy-based methods} guide generation with predefined or learned dialogue strategies, improving controllability and interpretability \cite{wang2019persuasion, liu2021towards, cheng2022improving, zhang2024escot, chang2024injectingsalespersonsdialoguestrategies}. 
\textit{Knowledge-enhanced methods} incorporate external knowledge or structured memory to improve contextual understanding and response quality \cite{sabour2022cem, deng2023knowledge, jia2023knowledge, liu2024improving}. 
\textit{Multi-agent systems} decompose dialogue into multiple coordinated agents, enabling more complex reasoning and collaborative decision-making \cite{cheng2024cooper, han2025tomaptrainingopponentawarellm, shi2025toward}. 

However, most existing approaches lack explicit modeling of users’ mental states, limiting their interpretability and long-term effectiveness in persuasive interactions. In contrast, our framework explicitly models the dependencies among these mental states in a stepwise manner, enabling more robust and interpretable inference to guide persuasive strategy selection.

\subsection{Theory of Mind in AI}
Theory of Mind refers to the ability to infer one’s own and others’ mental states. In the era of large language models, a variety of benchmarks have been proposed to evaluate ToM capabilities across diverse scenarios, including embodied environments, negotiation settings, narrative understanding, and persuasive interactions \cite{hou2410entering, chan2024negotiationtom, shinoda2025tomato, gandhi2023understanding, yu2025persuasivetom}. Existing evidence suggests that current LLMs still exhibit limited stability and reliability in mental-state inference, partly due to insufficient prior ToM knowledge encoded in these models.

To enhance ToM reasoning in LLMs, existing approaches can be broadly categorized into three lines. 
\textit{Prompting-based methods} improve ToM through inference-time reasoning scaffolds, such as perspective-taking prompts, role conditioning, and hypothesis-driven reasoning \cite{wilf2024think, shinoda2025let, huang2024notion, cross2024hypothetical}. 
\textit{Structured and neuro-symbolic approaches} explicitly model belief and goal inference by integrating symbolic representations or probabilistic reasoning frameworks \cite{sclar2023minding, ying2023neuro}. 
\textit{Multi-agent cognitive frameworks} decompose ToM reasoning into multiple interacting components to simulate social cognition processes \cite{zhang2025metamind}. 

Despite these advances, most existing methods primarily focus on reasoning over abstract or static states, and remain insufficient for fine-grained modeling of users’ dynamic mental states in real-world social interaction scenarios. Our approach addresses these limitations by combining stepwise backward inference over BDI states with knowledge-enhanced reasoning, enabling more accurate, consistent, and interpretable ToM inference in multi-turn persuasive dialogues.


\section{ToM based Persuasive Dialogue}\label{sec:tpmpd}
In a persuasive task, the persuader aims to influence the user’s attitude toward a persuasion target through appropriate persuasive strategies, ultimately inducing behavioral change. Achieving this goal requires the persuader to infer the persuadee’s latent mental states from the dialogue history and use these inferences to guide subsequent decisions. Accordingly, we incorporate the BDI model into persuasive dialogue and define the \textbf{ToM-PD} task.

This section introduces: (1) the formulation of the ToM-PD task, encompassing mental state inference and strategy prediction (Section~\ref{sec:tompd:task_definition}); and (2) a reverse and stepwise mental state inference procedure that treats human utterances as observable actions and sequentially infers intention, desire, belief and strategies (Section~\ref{sec:tompd:framework}).

\subsection{Task Formulation} \label{sec:tompd:task_definition}
The ToM-PD task aims to generate strategy-guided responses that effectively influence the persuadee’s mental state based on the dialogue history.

At dialogue turn t, the history is defined as $h_t = \{u_1, a_1, \dots, u_t\}$, where $u_k$ is the persuadee’s utterance and $a_k$ is the persuader’s response generated under strategy $s_k$. Given both $h_t$ and the inferred mental state $S_{\text{ToM}}^t = (i_t, d_t, b_t)$, representing intention, desire, and belief, the persuader predicts the next persuasive strategy $s_t$. The task is considered successful at turn t if the persuadee’s desire $d_{t+1}$ shifts toward the persuasion target.

\subsection{Reverse Mental State Inference} \label{sec:tompd:framework}
In traditional self  BDI state evolution process, psychological states are modeled through a forward process: beliefs give rise to desires, desires form intentions, and intentions lead to actions (Figure~\ref{fig:fig1}, left). In persuasive dialogue, the persuadee’s utterances are observable, requiring backward inference to recover underlying mental states (Figure~\ref{fig:fig1}, right).

As shown in figure~\ref{fig:fig1} (right), we propose a \emph{reverse mental state inference procedure} for ToM-PD. Starting from the dialogue history $h_t$, the persuadee’s mental states are inferred sequentially as:

\vspace{-5mm}
\begin{align}
    i_t &= f_{\text{intention}}(h_t), \\
    d_t &= f_{\text{desire}}(i_t), \\
    b_t &= f_{\text{belief}}(i_t, d_t),
\end{align}
\vspace{-5mm}

\noindent where $f_{\text{intention}}$, $ f_{\text{desire}}$, and $f_{\text{belief}}$ denote the inference functions for intention, desire, and belief, respectively.

Finally, the persuader selects the next strategy $s_t$ based on the inferred mental state:

\vspace{-3mm}
\begin{align}
    s_t = f_{\text{strategy}}(S_{\text{ToM}}^t),
\end{align}
\vspace{-5mm}

\noindent where $f_{\text{strategy}}$ denotes the selection function for strategy.

\section{ToM-BPD Dataset} \label{sec:tombpd}
PersuasiveToM is a multi-turn dialogue benchmark that spans diverse persuasive scenarios, designed to systematically evaluate the Theory-of-Mind reasoning capabilities of large language models in persuasive interactions \cite{yu2025persuasivetom}. Building upon this benchmark, we further construct the ToM-BPD dataset, which provides fine-grained annotations of the persuadee’s desire and belief states, as well as the strategies employed by the persuader, thereby capturing the key cognitive components underlying persuasive dialogue. To ensure data quality and annotation consistency, we develop a comprehensive data construction pipeline, comprising detailed annotation guidelines, annotator qualification tests, and multiple rounds of automated and human verification.

This section introduces: (1) the annotation framework of ToM-BPD, including the annotation schema and annotation procedure (Section~\ref{sec:tombpd:annotation}); and (2) the quality control process, detailing the specific measures adopted as well as the specific content of the designed tutorial (Section~\ref{sec:tombpd:quality}).

\subsection{Annotation}\label{sec:tombpd:annotation}
\subsubsection{Annotation Schema} \label{sec:tombpd:annotation:schema}
The primary annotation fields include the persuader’s strategy, as well as the persuadee’s desire and belief at each dialogue turn.

\noindent\textbf{Strategy Construction.} 
Based on the Elaboration Likelihood Model (ELM) \cite{petty1986elaboration} and Motivational Interviewing \cite{miller2013motivational}, we categorize persuasive strategies into three groups: socio-emotional strategies that foster empathy and strengthen rapport, cognitive strategies that influence reasoning through arguments and examples, and interactive strategies that facilitate dialogue and reduce resistance. Following prior work \cite{chen2025future}, we further refine these categories into nine fine-grained techniques. The taxonomy is summarized in Table~\ref{tab:strategy_taxonomy}, and detailed definitions with examples are provided in Table~\ref{tab:strategy_ToMBPD}.

\begin{table}[ht]
\centering
\caption{Taxonomy of persuasive strategies.}
\label{tab:strategy_taxonomy}
\renewcommand{\arraystretch}{1.2}
\begin{tabularx}{0.8\linewidth}{|>{\centering\arraybackslash}m{2.5cm}|>{\centering\arraybackslash}X|}
\hline
\textbf{Category} & \textbf{Technique}  \\
\hline
\multirow{3}{*}{\textbf{Socio-emotional}} 
& Affirmation and Reassurance \\
& Reflection of Feelings \\
& Personal Story \\
\hline
\multirow{5}{*}{\textbf{Cognitive}} 
& Expression of Views \\
& Enhancement of Views \\
& Logical Appeal \\
& Giving Examples \\
& Supplying Information \\
\hline
\multirow{1}{*}{\textbf{Interactive}}
& Task Inquiry \\
\hline

\end{tabularx}
\end{table}

\begin{table*}[htbp]
\centering
\caption{The strategies and their corresponding definitions in the ToM-BPD dataset}
\label{tab:strategy_ToMBPD}
\begin{tabularx}{0.97\textwidth}{|c| X| X|}
\hline
\multicolumn{1}{|c|}{\textbf{Strategy}} & 
\multicolumn{1}{c|}{\textbf{Definition}} & 
\multicolumn{1}{c|}{\textbf{Example}} \\
\hline
\textbf{Affirmation and reassurance} & 
refers to the strategy where the persuader validates the persuadee’s feelings, acknowledges difficulties, or encourages their sense of capability. & 
``I understand this is difficult, but I believe you can handle it." \\
\hline
\textbf{Reflection of feelings} & 
refers to the strategy where the persuader reflects, paraphrases, or interprets the persuadee’s emotional state to show understanding. & 
``It sounds like you're feeling overwhelmed right now." \\
\hline
\textbf{Personal story} & 
refers to the strategy where the persuader shares a personal experience or anecdote to enhance emotional resonance. & 
``I had a similar experience when I faced this situation last year." \\
\hline
\textbf{Expression of view} & 
refers to the strategy where the persuader expresses a personal standpoint, belief, or evaluation without necessarily providing reasoning. & 
``I think this option is better for you." \\
\hline
\textbf{Enhancement of view} & 
refers to the strategy where the persuader strengthens or intensifies a previously expressed stance through emphasis or elaboration. & 
``This is definitely the best choice for you, especially considering your goals." \\
\hline
\textbf{Logical appeal} & 
refers to the strategy where the persuader uses reasoning, cause-effect logic, or explicit argument structure. & 
``If you choose this option, you will therefore save both time and money." \\
\hline
\textbf{Supplying information} & 
refers to the strategy where the persuader provides factual information, general knowledge, or relevant advice. & 
``This program typically takes about six months to complete." \\
\hline
\textbf{Giving example} & 
refers to the strategy where a concrete instance or case is provided to support a point. & 
``For example, many students improved their performance using this method." \\
\hline
\textbf{Task inquiry} & 
refers to the strategy where the persuader asks an open-ended or exploratory question to understand the persuadee’s concerns or motivations. & 
``What concerns you most about this option?" \\

\hline
\end{tabularx}
\end{table*}

\noindent\textbf{Desire.} In persuasive dialogues, desire often reflects the user’s attitude toward the persuasion target (e.g., willingness to engage in an activity such as outdoor exercise). Previous work has represented user attitudes with discrete values \cite{deng2023plug,zhao2025dream,ma2026neuro,yang2026metro}. Following \cite{yang2026metro}, we operationalize desire as a discrete variable with three values, ${-1, 0, 1}$, corresponding to negative, hesitant, and positive attitudes toward the persuasion target. This representation indicates whether the user is currently inclined to reject, hesitate, or accept the persuasion goal, which is further used to determine the success of the persuasion task.

\noindent\textbf{Belief.} Beliefs are represented as short declarative statements reflecting the persuadee’s views, preferences, or comparisons related to the persuasion target. They are categorized as \textit{positive} (\textit{e.g., perceiving the target as interesting}) or \textit{negative} (\textit{e.g., concerns about potential risks}). A single dialogue turn may contain one or multiple such statements.

\subsubsection{Annotation Procedure} \label{sec:tombpd:annotation:procedure}
We adopt a semi-automatic annotation framework that combines automated pre-annotation with multi-round human verification. 

\noindent \textbf{Preprocessing.} Prior to annotation, we preprocess the dataset by removing dialogues involving more than two participants.

\noindent \textbf{Automatic Pre-annotation.} To improve annotation consistency and reduce human workload, we employ LLMs to perform automatic pre-annotation in the first annotation round. The prompts used for annotation are provided in Section~\ref{sec:app:prompt:annotation}.

\noindent \textbf{Annotation Correction.}
Human annotators review and verify the automatic annotations. During this stage, annotators are instructed to flag ambiguous or difficult-to-judge dialogue segments, as well as samples with inconsistent labels within the same dialogue. After each annotation round, the involved annotators are organized to discuss the flagged samples and reach consensus by articulating their judgments; samples for which consensus cannot be reached are discarded. In the end, we remove 21 dialogues from the original 525 dialogues in the PersuasiveToM dataset.

\subsection{Quality Control}\label{sec:tombpd:quality}
To ensure annotation consistency and data quality, we adopt the following measures: (1) we design systematic annotation tutorials for annotators; (2) we design qualification tests for annotators; (3) we integrate automated annotation with human annotation and implement a multi-stage quality control process(Section~\ref{sec:tombpd:annotation:procedure}). 

\noindent \textbf{Tutorial.} We design a structured training program for volunteer annotators. The training materials cover fundamental theoretical knowledge of ToM, following the definitions and formalizations in~\cite{yu2025persuasivetom}, as well as foundational theories of strategy-based persuasive dialogue based on~\cite{chen2025future}. In addition, the tutorial introduces core annotation principles for ToM mental states, and provides formal definitions of persuasive strategies (Table~\ref{tab:strategy_ToMBPD}) together with illustrative examples.

\noindent \textbf{Annotation Principles for ToM States.} 
Let $n$ denote the dialogue turn number. For the first turn ($n=1$), annotation is performed at the sentence level. If the persuadee expresses appreciation for the persuasive target, such as ``\textit{This sounds interesting}'' or ``\textit{This seems promising},'' the sentence is labeled as a positive sentence. Conversely, if the persuadee expresses concerns or uncertainty (e.g., ``\textit{I am not sure...}''), or explicitly supports an event contrary to the persuasive goal, the sentence is labeled as a negative sentence. Neutral sentences are ignored when determining the desire label.

At the beginning of the dialogue, if the persuadee’s current utterance contains only positive sentences, the \textit{desire} is labeled as $1$; if it contains both positive and negative sentences, it is labeled as $0$; and if it contains only negative sentences, it is labeled as $-1$. The extracted positive and negative sentences are recorded as natural language descriptions of beliefs, for example: (Positive: \textit{attending the activity is interesting}; Negative: \textit{uncertain about the benefit of the activity}).

For subsequent turns ($n>1$), the negative sentences from the previous turn are examined. A negative sentence is considered resolved if it is explicitly addressed or mitigated in the subsequent dialogue. If all negative sentences are resolved, annotation proceeds following the same principles as in the first turn. Otherwise, unresolved negative sentences are retained, and the current turn is annotated by considering both newly expressed positive and negative sentences together with the unresolved ones. The resulting positive and negative sentences are similarly recorded as belief descriptions.

\noindent \textbf{Examination.} 
We designed a qualification test comprising 25 labeled dialogue turns to evaluate volunteers’ understanding of the annotation concepts and their ability to correctly apply the labeling standards. Specifically, volunteers were required to annotate, based on the current dialogue history and its progression, the persuader’s strategy as well as the persuadee’s desire and belief. We then calculated the average annotation accuracy across these three dimensions for each annotator. Annotators achieving an accuracy above 90\% were considered qualified for the annotation task. Ultimately, this test resulted in the disqualification of 8 out of 23 participants.

\subsection{Data Characteristics}\label{sec:tombpd:characteristics}
\noindent \textbf{Overall Statistics.} Table~\ref{tab:dataset_comparison} reports the statistics of ToM-BPD. The dataset contains 504 dialogues and 3,926 utterances, with an average of 7.79 turns per dialogue. The average utterance length differs across roles, with 38.62 tokens for the persuader and 19.67 tokens for the persuadee. Each turn contains 1.89 belief annotations on average, suggesting that persuadee responses are often driven by multiple underlying beliefs. The average desire score is -0.17, indicating that most dialogues begin with a certain degree of resistance toward the persuasion target.

\begin{table}[ht]
\centering
\caption{Overall statistics of ToM-BPD.}
\label{tab:dataset_comparison}
\resizebox{0.8\linewidth}{!}{%
\begin{tabular}{|l|c|}
\hline
\textbf{Statistic} & \textbf{ToM-BPD}  \\
\hline
Total dialogues & 504  \\
Total utterances & 3926 \\
Avg. dialogue length (turns) & 7.79 \\
\hline
Avg. utterance length (persuader) & 38.62 \\
Avg. utterance length (persuadee) & 19.67 \\
\hline
Avg. desire & -0.17  \\
Avg. belief number per turn & 1.89  \\
\hline
\end{tabular}%
}
\end{table}

\noindent \textbf{Percentages of Strategies.} Table~\ref{tab:strategy_hierarchical_bpd} presents the distribution of persuasive strategies in ToM-BPD. Cognitive strategies dominate the dataset (60.84\%), followed by socio-emotional strategies (34.36\%), while interactive strategies account for a relatively small proportion (4.79\%). In conjunction with the functional definitions in Table~\ref{tab:strategy_ToMBPD}, this distribution suggests that persuaders in ToM-BPD primarily influence users through cognitive strategies, complemented by socio-emotional alignment, with comparatively limited use of interaction-driven strategies.

\begin{table}[ht]
\centering
\caption{Distribution of strategies in ToM-BPD.}
\label{tab:strategy_hierarchical_bpd}
\resizebox{0.95\linewidth}{!}{%
\begin{tabular}{|c|c|c|c|}
\hline
\textbf{Category} & \textbf{Strategy} & \textbf{Count} & \textbf{Percentage} \\
\hline
\multirow{3}{*}{\textbf{Socio-emotional}} 
& Affirmation and Reassurance & 576 & 13.79\% \\
& Reflection of Feelings      & 698 & 16.72\% \\
& Personal Story              & 161 & 3.86\% \\
\hline
\multirow{5}{*}{\textbf{Cognitive}} 
& Expression of Views         & 1241 & 29.74\% \\
& Enhancement of Views        & 155  & 3.71\% \\
& Logical Appeal              & 416  & 9.97\% \\
& Giving Examples             & 224  & 5.36\% \\
& Supplying Information       & 505  & 12.09\% \\
\hline
\multirow{1}{*}{\textbf{Interactive}} 
& Task Inquiry                & 200  & 4.79\% \\

\hline
\end{tabular}%
}
\end{table}

\begin{figure}[ht]
  \centering
  \includegraphics[width=\linewidth]{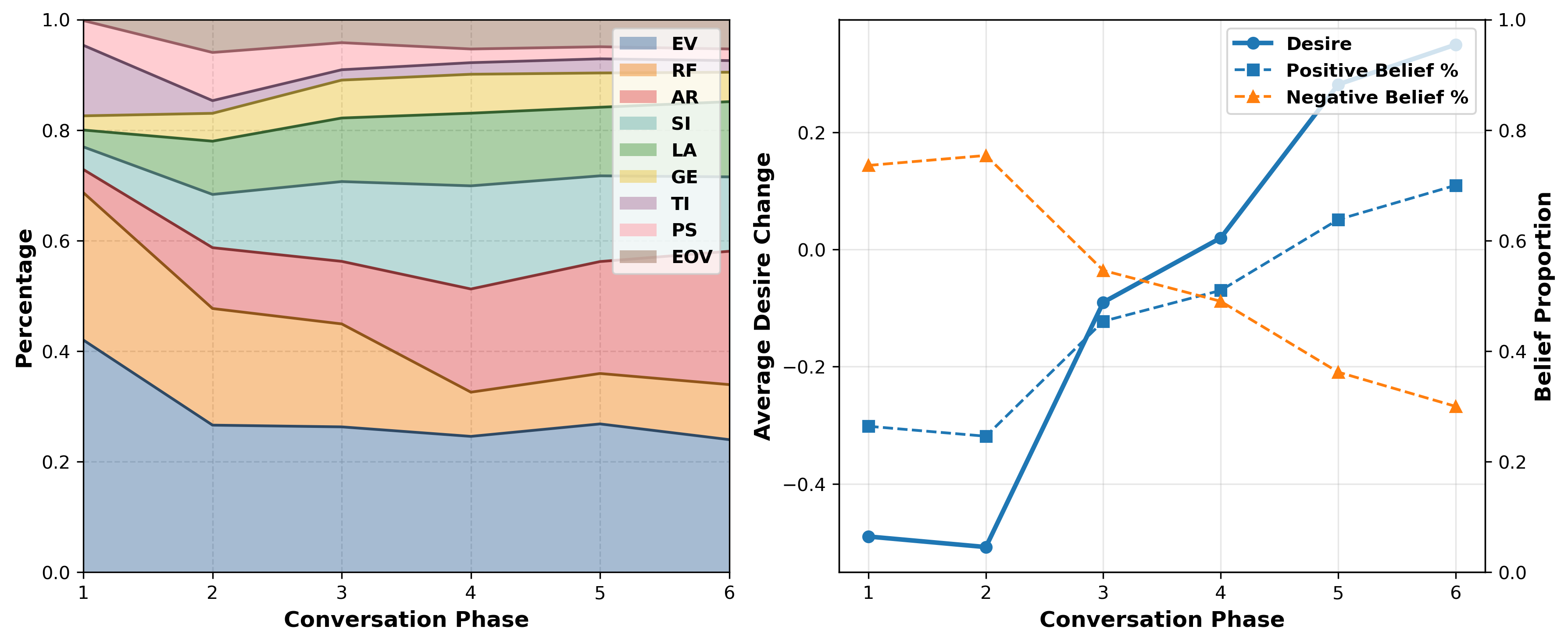}
  \caption{Overall analysis of dialogue strategies, desire, and belief dynamics across conversation phases. 
  The left panel shows the distribution of strategies at different stages of the dialogue, while the right panel illustrates the trajectory of desire and the proportions of positive and negative beliefs.}
  \label{fig:dual_strategy_analysis}
\end{figure}

\noindent \textbf{Strategy Distribution.} We analyzed the distribution of persuasive strategies across different stages of the conversation. Following the design in~\cite{liu2021towards}, we compute the proportion of each strategy within each interval. The resulting strategy distributions are plotted at the six representative progress points and connected to illustrate the temporal evolution of persuasive strategy usage throughout the conversation. As shown in Figure~\ref{fig:dual_strategy_analysis}, In early stages, strategies focus on empathizing and introducing the target. In later stages, diverse strategies guide changes in the persuadee’s desire.

\noindent \textbf{Desire Trajectory.} Figure~\ref{fig:dual_strategy_analysis} (left) shows the average desire trend of the persuadee. Desire starts low at target introduction and gradually rises as the dialogue progresses.

\noindent \textbf{Belief Characteristics.} Figure~\ref{fig:dual_strategy_analysis} (right) also shows the proportions of positive and negative beliefs over the dialogue. These proportions closely follow the desire trajectory, supporting the rationale for modeling beliefs in \textsc{ToM-BPD}.

\section{Think Thrice Before You Speak} \label{sec:method}
\textbf{TTBYS} enhances the reasoning capabilities of LLMs on ToM-related tasks by leveraging prior ToM experiences. Each experience is represented as a quintuple consisting of dialogue history, intention, desire, belief, and strategy (Section~\ref{sec:method:experience}). 

\begin{figure*}[th]
  \centering
  \includegraphics[width=\linewidth]{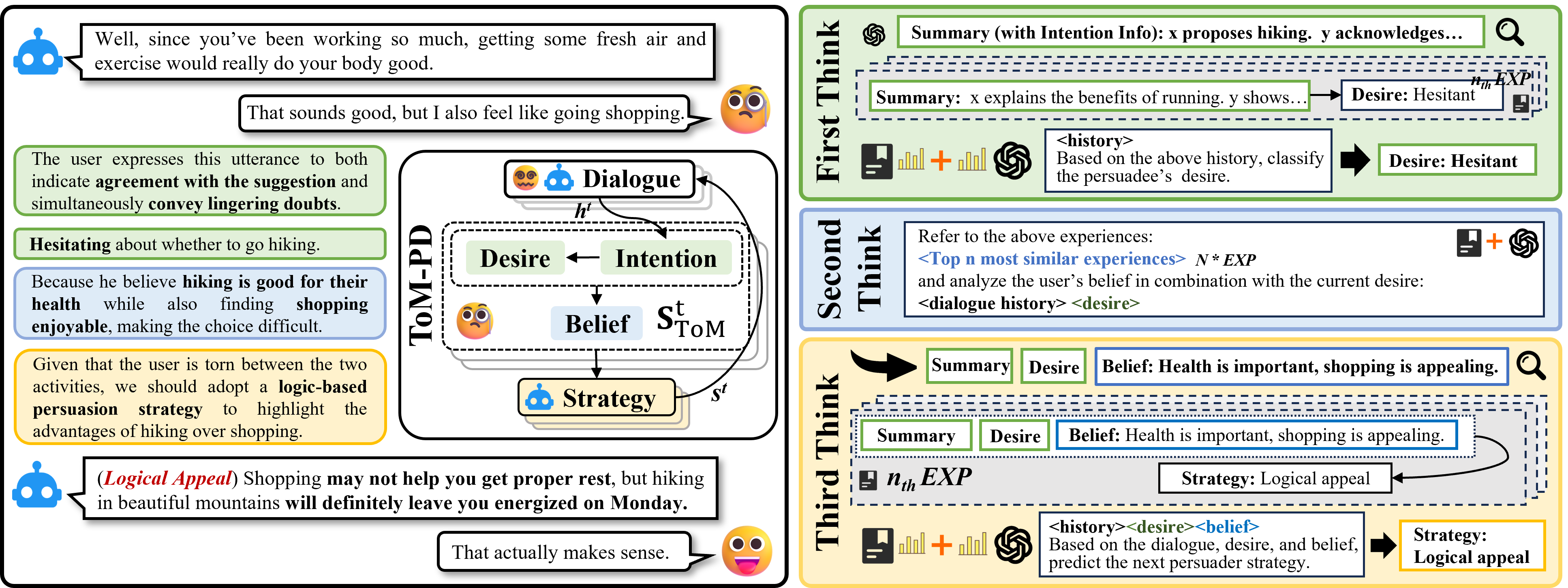}
  \caption{Overview of the \textbf{ToM-PD} (left) and the \textbf{TTBYS} framework (right). In the ToM-PD task, the persuader sequentially infers the persuadee’s mental states, including intention, desire, and belief, from the dialogue history, and subsequently selects an appropriate persuasive strategy based on the inferred states. TTBYS operationalizes this process through three explicit reasoning steps, each corresponding to one stage of mental state inference.}
  \label{fig:overview}
\end{figure*}

The overall reasoning process is illustrated in Figure~\ref{fig:overview}. First, relevant experiences are retrieved based on the dialogue summary to construct a probability distribution for implicitly guiding the LLM’s desire inference (Section~\ref{sec:method:first_think}). Second, conditioned on the dialogue summary and the inferred desire, retrieved experiences are incorporated into the LLM context to explicitly facilitate belief generation (Section~\ref{sec:method:second_think}). Finally, experiences are jointly retrieved using the dialogue summary, desire, and belief to implicitly enhance strategy prediction by the LLM (Section~\ref{sec:method:third_think}).

\subsection{ToM-PD Experience}\label{sec:method:experience}
In human persuasive interactions, individuals intuitively retrieve analogous past scenarios from memory, infer the interlocutor's latent mental states, and select adaptive strategies. Inspired by this cognitive process, we formalize the \textbf{ToM-based Persuasive Dialogue Experience (ToM-PD Experience)} as a structured knowledge unit capturing the mapping among dialogue context, the persuadee's mental states, and corresponding persuasive strategies.

To operationalize this experience, we introduce a \emph{dialogue summary} mechanism to mitigate LLMs' limitations in intention inference. Given the current dialogue turn $(u_t, a_t)$, we abstract it into a summary $i$, which preserves the key information relevant to the persuadee's intention and prevents error accumulation caused by reasoning from incorrectly inferred intentions. The prompt for generation of dialogue summary is as follows.

\begin{promptbox}{Dialogue Summary Generation Prompt}
\small
\vspace{-3mm}
    You are an annotation system for Theory-of-Mind dialogue analysis. \\
    Your task is to generate a concise summary for a persuasive dialogue, where the roles include the persuader (x) and the persuadee (y). \\
    The summary should include:
    (1) the main persuasion strategy used by x, and
    (2) the final attitude or response tendency of y. \\
    Rules
    - Focus on high-level semantics; do not repeat specific dialogue content.
    - Do not directly copy sentences from the dialogue.
    - The summary should be limited to one or two sentences.
    - Use “x” to refer to the persuader and “y” to refer to the persuadee.
    \texttt{<an example>}
    \texttt{<dialogue history>} \\
\vspace{-5mm}
\end{promptbox}

\begin{figure}[!ht]
  \centering
  \includegraphics[width=0.85\linewidth]{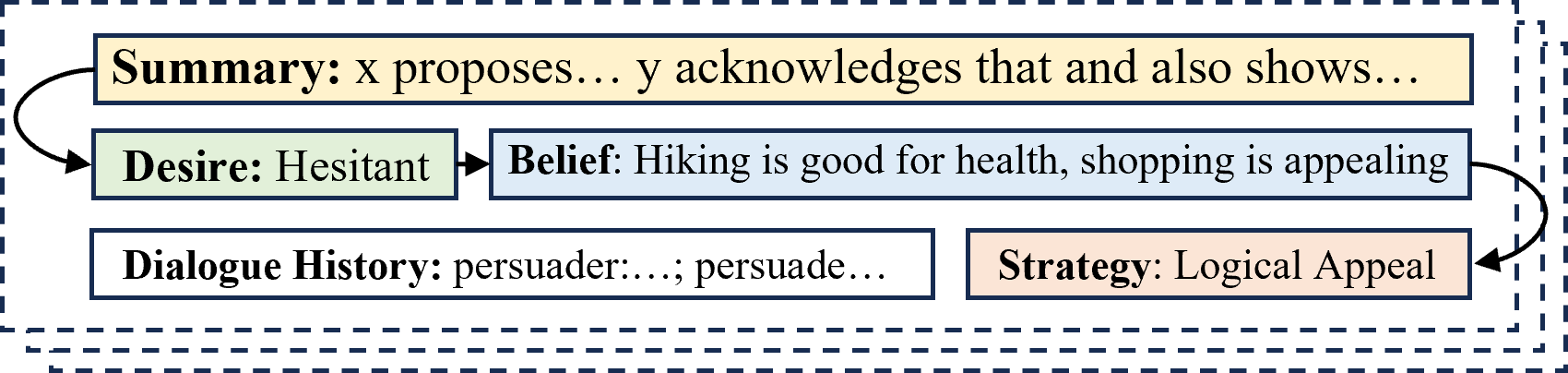}
  \caption{An example of a ToM-PD Experience.}
  \label{figure:tom_experience}
\end{figure}

We further model a ToM-PD Experience as a quintuple $(h_t, i, d, b, s)$, where $h_t$ is the dialogue history, $i$ is the summary encapsulating the persuadee's intention, $d$ and $b$ denote the persuadee's desire and belief, respectively, and $s$ represents the strategy corresponding to the current ToM state $(i, d, b)$ (Figure~\ref{figure:tom_experience}). The knowledge base consists of multiple such quintuples, where $i$ is generated directly by the LLM, and the remaining fields are derived from the ToM-BPD dataset. In practice, a single conversation is decomposed into multiple experiences, each treated as an independent knowledge unit for experience retrieval and reasoning enhancement. 

\subsection{First Think: Desire Inference}\label{sec:method:first_think}
To leverage ToM experiences for deliberative judgment, we first summarize the current dialogue turn $(u_t, a_t)$ as a \textit{dialogue summary} $i_t$. This summary captures the key observable behaviors and the inferred intention of the persuadee at this stage. Using $i_t$ as a query, we retrieve the top-$N$ most semantically similar historical experiences from the ToM knowledge base. Based on the desire annotations associated with these retrieved experiences, we construct an experience-driven desire distribution as a \textbf{implicit knowledge-based enhancement}:

\vspace{-3mm}
\begin{equation}
P_{\text{ex}}(D)\! = \!\left(\!P_{\text{ex}}(d^{(-1)}), \!P_{\text{ex}}(d^{(0)}),\! P_{\text{ex}}(d^{(1)})\right),
\end{equation}

\noindent where $D = \{-1, 0, 1\}$ represents unwillingness, hesitation, and willingness toward the persuasion target, respectively, and $P_{\text{ex}}(d)$ denotes the normalized proportion of retrieved experiences with desire $d$, satisfying $\sum_{d \in D} P_{\text{ex}}(d) = 1$.

To simulate fast, intuitive human judgments, we also extract the conditional probability distribution over desire states from the LLM’s first generated token when predicting desire:

\vspace{-3mm}
\begin{equation}
\!P_{\text{ag}}(\!D)\! =\!\left(\!P_{\text{ag}}(\!d^{(-1)}\!), P_{\text{ag}}(\!d^{(0)}), P_{\text{ag}}(\!d^{(1)}\!)\!\right),
\end{equation}

\noindent where $P_{\text{ag}}(d)$ denotes the LLM-assigned probability of desire $d$ given the current dialogue context, with $\sum_{d \in D} P_{\text{ag}}(d) = 1$.

Finally, to balance the reliability of experience-based reasoning and the flexibility of LLM-driven intuition, we linearly fuse the two distributions. The predicted desire $d_t^*$ is obtained by maximizing the fused posterior:

\vspace{-3mm}
\begin{equation}
d_t^*\!=\!\arg\max_{d \in D}\left(\!\alpha \!\cdot\! P_{\text{ag}}(\!d_t)\!+\!(\!1-\alpha) \!\cdot\! P_{\text{ex}}(\!d_t)\right),
\end{equation}

\noindent where $\alpha \in [0,1]$ controls the relative contribution of the LLM’s intuitive judgment and the experience-based inference. The prompt for desire prediction is presented as follows:

\begin{promptbox}{Prompt for Desire Prediction}
\small
\vspace{-3mm}
    Current conversation: \texttt{<dialogue history>} \\
    Classify the persuadee's desire based on the above conversation.\\
    Choose exactly one option: \\
    A. Unwilling \\
    B. Uncertain \\
    C. Willing \\
    Answer with ONLY A, B, or C. \\
    Do not output anything else.
    \vspace{-3mm}
\end{promptbox}

\subsection{Second Think: Belief Inference}\label{sec:method:second_think}
We next infer the persuadee’s belief $b$, which causally underlies the inferred desire $d^*$. To this end, we use the dialogue summary $i$ corresponding to $d^*$ as a retrieval query. Based on this query, we retrieve the top-$N$ most relevant experiences from the ToM knowledge base, which contain dialogue-history–belief information aligned with the current dialogue context and the inferred desire. These retrieved experiences are then directly incorporated into the LLM’s context as an \textbf{explicit knowledge-based enhancement}, enabling the model to generate the belief $b^*$ for the current dialogue $h$. Formally, this process is expressed as:

\vspace{-2mm}
\begin{equation}
b_t^* = \text{LLM}(\xi, \text{exp}, d_t^*),
\end{equation}

\noindent where \(\text{LLM}(\cdot)\) denotes an LLM, \(\xi\) represents the task description, and \(\text{exp}\) denotes the experiences. In supplementary materials, we provide concrete examples of using these experiences for belief inference. The prompt we used for belief prediction is presented as follows:

\begin{promptbox}{Prompt for Belief Prediction}
\small
\vspace{-3mm}
Relevant Experience: \texttt{<top relevant experience>} \\
Infer the persuadee's belief in the current conversation context based on the prediction method in relevant experiences. \\
Current conversation: \texttt{<dialogue history>} \\
Desire level: \texttt{<desire>} \\
Generate a single-line natural language description of the persuadee's belief.
\vspace{-3mm}
\end{promptbox}

\subsection{Third Think: Strategy Prediction}\label{sec:method:third_think}

After obtaining the complete ToM state $(i_t, d^*, b^*)$, we aim to select the optimal discrete persuasive strategy $s_t$ to guide the persuadee’s desire $d^*$ toward accepting the persuasion target. To this end, we perform a joint retrieval using the dialogue summary $i$ and the inferred belief $b^*$ as the query, assigning equal weights to each. This retrieves the top-$N$ experiences most relevant to the current persuasive scenario and the persuadee’s mental state. 

Following the approach in \textbf{First Think}, we construct both an experience-driven strategy distribution and an LLM-driven probability distribution. We then introduce a coefficient \( \beta \) to fuse the two distributions, resulting in an optimal strategy that is both experience-consistent and intuition-aligned:

\vspace{-3mm}
\begin{equation}
s_t^*\! =\! \arg \max_{s \in \mathcal{S}}\left(\!\beta\! \cdot\! P_{\text{\!ag}}(\! S_{\text{ToM}}^t\!)\! + \!(\!1\!-\!\beta) \!\cdot \!P_{\text{\!ex}}(\! S_{\text{\!ToM}}^t\!) \!\right)\!. \end{equation}

\noindent where $P_{\text{ex}}(S_{\text{ToM}}^t)$ and $P_{\text{ag}}(S_{\text{ToM}}^t)$ denote the experience-driven and LLM-driven strategy distributions, respectively. In Section~\ref{sec:exp:static:ablation}, we systematically analyze the effects of coefficient $\alpha$ and coefficient $\beta$ on desire prediction and strategy prediction accuracy. And the prompt we used for strategy prediction is presented as follows:

\begin{promptbox}{Prompt for Strategy Prediction}
\small
\vspace{-3mm}
    Current conversation: \texttt{<dialogue history>} \\
    Desire level: \texttt{<desire>} \\
    Belief: \texttt{<belief>} \\
    Strategy definitions: \texttt{<Strategy definitions>} \\
    Based on the dialogue, desire, and belief, predict the next persuader strategy. \\
    Return ONLY ONE of the above single-letter labels: V, L, E, T, P, A, R, I, G. \\
    Do not output anything else.
    \vspace{-3mm}
\end{promptbox}


\section{Experiment} \label{sec:exp}
This section presents the experimental evaluation of TTBYS. Section~\ref{sec:exp:setup} describes the experimental setup, Section~\ref{sec:exp:static} presents the static evaluation, Section~\ref{sec:exp:inter} presents the interactive evaluation, Section~\ref{sec:exp:case} shows case studies, and Section~\ref{sec:exp:time} reports runtime statistics.

\subsection{Experimental Setup} \label{sec:exp:setup}
\noindent \textbf{Datasets}
We conduct experiments on the \textsc{ToM-BPD} dataset. The first 100 conversations with 399 turns, are used as the test set, while the remaining 404 conversations with 1,564 ToM experiences constitute the knowledge base.

\noindent \textbf{Baselines}
We compare the proposed method with two basic prompting techniques and two recent ToM-enhanced reasoning methods across six frontier large language models. The basic prompting techniques include vanilla zero-shot prompting and chain-of-thought (\textbf{CoT}) prompting \cite{kojima2022large,sabour2024emobench}. In addition, we consider two representative ToM-enhanced methods, \textsc{Hypothetical Minds} (\textbf{HM}) and \textsc{MetaMind} (\textbf{MM}), whose frameworks are adopted with prompt templates adapted to our task \cite{zhang2025metamind,cross2024hypothetical}. 

The evaluated models include three open-source models, LLaMA-3.1-8B-Instruct \cite{grattafiori2024llama}, Qwen-3-8B \cite{yang2025qwen3}, and Mixtral-7B-Instruct \cite{jiang2024mixtral}, as well as three closed-source models, Gemini-3-Pro, \textsc{GPT-4o-mini}, and \textsc{GPT-5}. Due to the lack of access to token-level probabilities (e.g., softmax outputs) required by our method, it is not applied to closed-source models in Section~\ref{sec:exp:static:main}.

\noindent \textbf{Implementation Details}
All models are evaluated with a temperature of 0.9, and the reported results are averaged over three runs. In the main experiments, the blending coefficients $\alpha$ and $\beta$ are set to 0.5 and 0.3, respectively. The number of retrieved experiences $N$ is set to 5 for First Think and Second Think, and 10 for Third Think. For experience retrieval, we employ the all-MiniLM-L6-v2 model as the sentence embedder and use cosine similarity as the metric for selection. Prompt templates are provided in Appendix~\ref{sec:app:prompt}. All experiments were conducted on a single 80GB A100 GPU.

\noindent \textbf{Evaluation Metrics}
We evaluate belief reasoning and strategy prediction using accuracy. As beliefs do not belong to a fixed label space, we employ a GPT-5–based prompt evaluation protocol for belief prediction. For each instance, a score of 1 is assigned if both the belief polarity (positive/negative) and its underlying reasons match the ground truth; a score of 0.5 is assigned if only the polarity is correct; otherwise, a score of 0 is given. Belief prediction accuracy is reported as the average score across all instances.

\subsection{Static Evaluation} \label{sec:exp:static}
This section presents the static experiments of TTBYS. Section~\ref{sec:exp:static:main} reports the comparison with baselines, Section~\ref{sec:exp:static:ablation} studies the effect of blending coefficients on prediction accuracy, Section~\ref{sec:exp:static:num_impact} analyzes the impact of the number of experience samples on prediction accuracy, and Section~\ref{sec:exp:static:abs_impact} investigates the effect of using dialogue summaries as retrieval targets.

\begin{table}[t]
\centering
\small
\caption{Prediction accuracy (\%) of desire, belief and strategy on the ToM-BPD dataset. 
Grey-shaded rows represent the proposed method, and bold values denote the best results under each metric.}
\resizebox{1.0\columnwidth}{!}{
\begin{tabular}{|l|c|c|c|}
\hline
\textbf{Model and Method} & \textbf{Desire Acc.} & \textbf{Belief Acc.} & \textbf{Strategy Acc.} \\
\hline
GPT-4o-mini           & 65.82$_{\pm0.60}$ & 21.45$_{\pm2.51}$ & 27.13$_{\pm0.72}$ \\
GPT-4o-mini + CoT     & 66.33$_{\pm0.54}$ & 23.41$_{\pm2.48}$ & 26.68$_{\pm0.65}$ \\
GPT-4o-mini + HM      & 64.87$_{\pm0.76}$ & 30.60$_{\pm2.48}$ & 27.72$_{\pm1.48}$ \\
GPT-4o-mini + MM      & 65.15$_{\pm0.85}$ & \textit{31.96$_{\pm2.38}$} & \textit{28.18$_{\pm1.25}$} \\
\hline
GPT-5                 & 71.62$_{\pm0.47}$ & 21.84$_{\pm2.36}$ & 22.81$_{\pm0.53}$ \\
GPT-5 + CoT           & 70.91$_{\pm0.38}$ & 22.68$_{\pm2.34}$ & 21.16$_{\pm0.44}$ \\
GPT-5 + HM            & 68.33$_{\pm0.87}$ & 29.60$_{\pm2.12}$ & 26.68$_{\pm1.32}$ \\
GPT-5 + MM            & 69.33$_{\pm0.91}$ & 31.74$_{\pm1.89}$ & 27.14$_{\pm1.29}$ \\
\hline
Gemini-3-pro-high     & \textit{72.26$_{\pm0.51}$} & 17.84$_{\pm2.49}$ & 20.67$_{\pm0.63}$ \\
Gemini-3-pro-high + CoT & 70.68$_{\pm0.46}$ & 18.34$_{\pm2.71}$ & 19.84$_{\pm0.55}$ \\
Gemini-3-pro-high + HM  & 68.84$_{\pm0.98}$ & 27.90$_{\pm2.36}$ & 23.68$_{\pm1.65}$ \\
Gemini-3-pro-high + MM  & 66.33$_{\pm1.07}$ & 29.28$_{\pm2.17}$ & 24.59$_{\pm1.58}$ \\
\hline
LLaMA-3.1-8B-Instruct  & 34.41$_{\pm1.21}$ & 25.67$_{\pm1.03}$ & 14.03$_{\pm1.58}$ \\
LLaMA-3.1-8B-Instruct + CoT & 36.25$_{\pm1.14}$ & 27.69$_{\pm1.31}$ & 16.80$_{\pm1.28}$ \\
LLaMA-3.1-8B-Instruct + HM  & 54.14$_{\pm0.91}$ & 30.80$_{\pm2.48}$ & 22.12$_{\pm1.73}$ \\
LLaMA-3.1-8B-Instruct + MM  & 66.33$_{\pm1.07}$ & 31.67$_{\pm1.87}$ & 22.63$_{\pm1.47}$ \\
\rowcolor{gray!15}
\textbf{LLaMA-3.1-8B-Instruct + ours} & 69.43$_{\pm0.65}$ & 43.62$_{\pm1.57}$ & 37.76$_{\pm0.73}$ \\
\hline
Qwen-3-8B             & 50.43$_{\pm1.07}$ & 38.64$_{\pm1.78}$ & 21.42$_{\pm0.91}$ \\
Qwen-3-8B + CoT       & 50.90$_{\pm1.01}$ & 39.42$_{\pm1.79}$ & 19.86$_{\pm0.97}$ \\
Qwen-3-8B + HM        & 65.84$_{\pm0.88}$ & 41.21$_{\pm2.48}$ & 23.68$_{\pm1.71}$ \\
Qwen-3-8B + MM        & 64.33$_{\pm1.28}$ & 42.05$_{\pm2.17}$ & 21.89$_{\pm1.68}$ \\
\rowcolor{gray!15}
\textbf{Qwen-3-8B + ours} & \textbf{72.82$_{\pm0.58}$} & \textbf{54.64$_{\pm2.46}$} & \textbf{39.78$_{\pm0.63}$} \\
\hline
Mixtral-7B-Instruct   & 45.64$_{\pm0.84}$ & 22.86$_{\pm1.93}$ & 16.41$_{\pm1.19}$ \\
Mixtral-7B-Instruct + CoT & 44.10$_{\pm0.97}$ & 24.13$_{\pm1.79}$ & 17.64$_{\pm1.04}$ \\
Mixtral-7B-Instruct + HM  & 60.22$_{\pm1.24}$ & 26.80$_{\pm1.49}$ & 17.84$_{\pm1.25}$ \\
Mixtral-7B-Instruct + MM  & 59.70$_{\pm1.21}$ & 29.06$_{\pm1.41}$ & 17.12$_{\pm1.54}$ \\
\rowcolor{gray!15}
\textbf{Mixtral-7B-Instruct + ours} & 71.44$_{\pm0.53}$ & 36.96$_{\pm2.47}$ & 36.79$_{\pm0.61}$ \\
\hline
\end{tabular}
}
\label{tab:main_result}
\end{table}

\subsubsection{Comparison with Baselines} \label{sec:exp:static:main}
Table~\ref{tab:main_result} shows the prediction accuracy of different models and methods on the ToM-BPD dataset across desire, belief, and strategy prediction.

Under basic prompting and CoT settings, closed-source models consistently outperform open-source models on desire and strategy prediction, with a particularly pronounced performance gap on the desire dimension. In contrast, open-source models exhibit limited ToM reasoning capabilities in these settings, indicating that relying solely on surface-level dialogue signals is insufficient for inferring latent mental states.

For the belief prediction task, all models show limited performance under basic prompting conditions, reflecting the inherent difficulty of belief inference. Although CoT prompting yields stable but modest improvements, its overall gains remain constrained. Existing enhancement methods, including \textsc{HM} and \textsc{MM}, significantly improve belief prediction by explicitly modeling mental states; however, their improvements on desire and strategy prediction are marginal. This suggests that the lack of prior knowledge about ToM reasoning substantially limits models’ inference performance.

In contrast, incorporating TTBYS leads to substantial and consistent performance gains across all three prediction dimensions for all open-source models. Notably, Qwen-3-8B + ours achieves the best overall performance on desire, belief, and strategy prediction, comprehensively outperforming both closed-source models and existing enhancement methods. Similar performance trends are also observed for other open-source backbones. Two cases are provided in the supplementary material.

Overall, TTBYS consistently improves ToM reasoning and strategy prediction across diverse open-source backbones, enabling them to match or surpass closed-source models through effective utilization of structured ToM experiences.

\begin{figure}[t]
    \centering
    \includegraphics[width=0.95\linewidth]{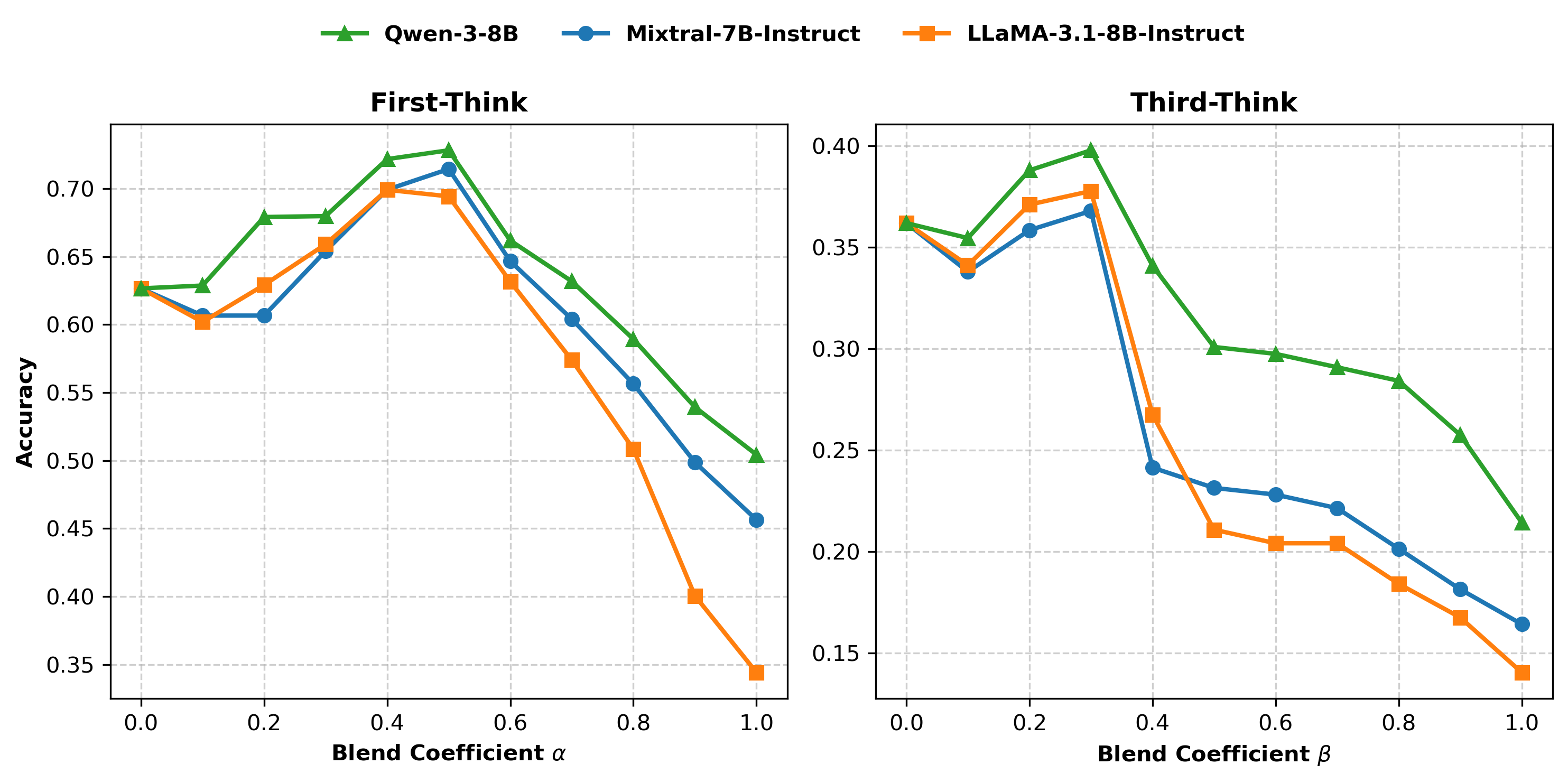}
    \caption{Impact of the blending coefficients on prediction accuracy. Left: first-think (desire prediction) with varying $\alpha$. Right: third-think (strategy prediction) with varying $\beta$. }
    \label{fig:ablation}
\end{figure}

\subsubsection{Analysis of Blending Coefficients} \label{sec:exp:static:ablation}
To evaluate the contributions of the LLM-based and experience-based components, we vary the blending coefficients $\alpha$ (first-think) and $\beta$ (third-think) across three backbone models: Qwen-3-8B, LLaMA-3.1-8B-Instruct, and Mixtral-7B-Instruct. Setting $\alpha$ or $\beta$ to 0 relies solely on expert experience, whereas values of 1 correspond to using only the LLM. Desire prediction is analyzed in first-think, and strategy prediction in third-think. The knowledge base setup follows Section~\ref{sec:exp:setup}.

As illustrated in Figure~\ref{fig:ablation}, integrating ToM-PD experience consistently boosts accuracy across all models. Peak performance for first-think is observed when $\alpha$ is around 0.4–0.6, while third-think achieves its highest accuracy at a smaller $\beta$ value of 0.2–0.4. This indicates that during strategy reasoning, the models rely more heavily on experiences rather than solely on the LLM. Moreover, all three models reach their peaks at similar blending values in both first-think and third-think, suggesting that different backbones exhibit comparable reasoning distributions when performing ToM inference tasks. These results demonstrate that a balanced combination of experience-based and LLM-based signals outperforms reliance on either component alone.

\begin{figure}[t]
    \centering
    \includegraphics[width=0.95\linewidth]{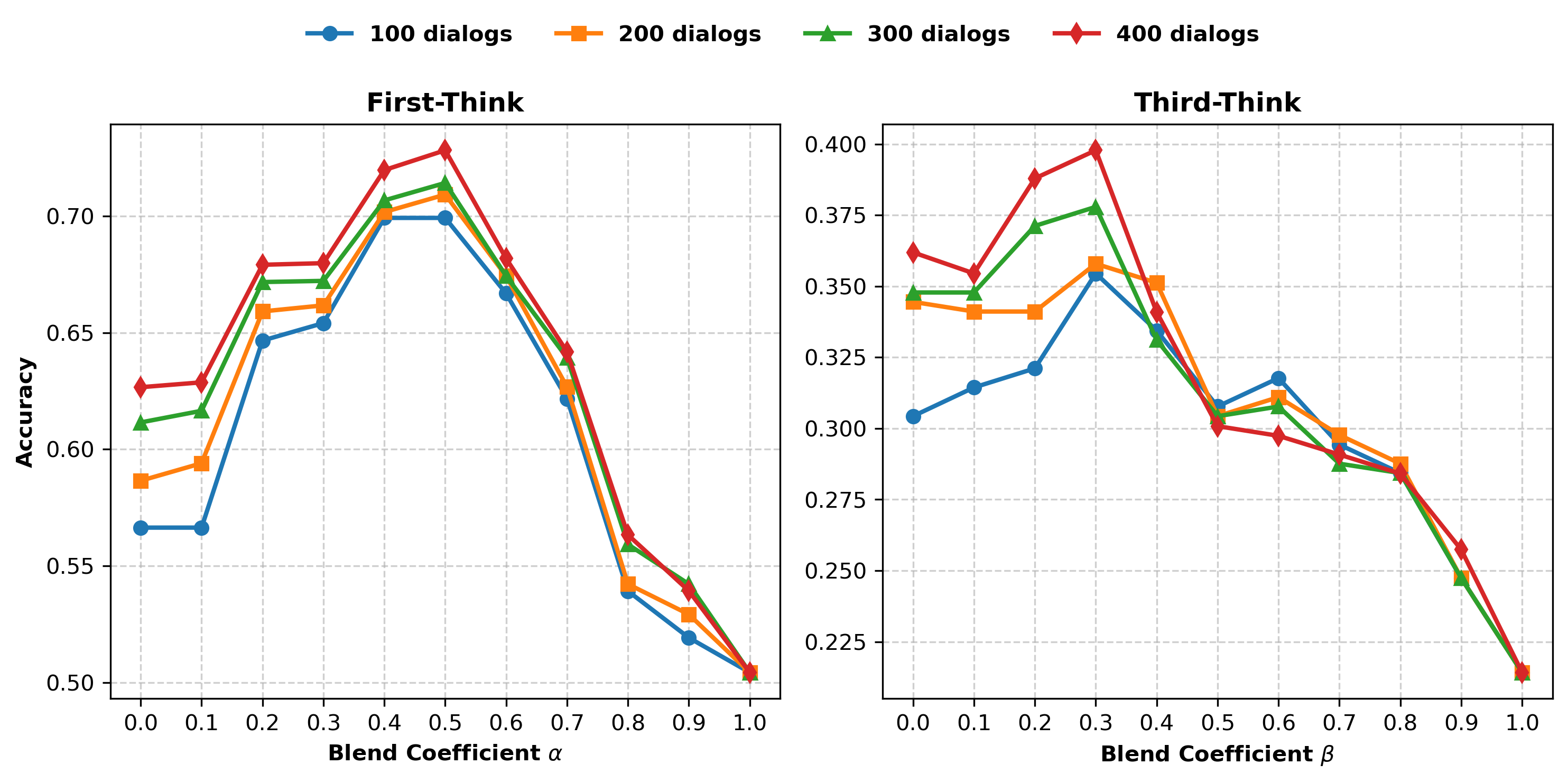}
    \caption{Impact of experience quantity on desire and strategy prediction performance in the first-think and third-think stages. Left: first-think  with varying $\alpha$. Right: third-think with varying $\beta$.}
    \label{fig:impact_of_num}
\end{figure}

\begin{table}[t]
    \centering
    \caption{Effect of experience quantity on belief prediction performance in the second-think stage. Accuracy (\%) is reported for Belief prediction.}
    \resizebox{0.9\linewidth}{!}{%
    \begin{tabular}{|c|c|c|c|c|}
    \hline
    \textbf{Num} 
    & \textbf{100} 
    & \textbf{200} 
    & \textbf{300}
    & \textbf{400}\\
    \hline
    \textbf{Belief Acc.}  
    & 51.75$_{\pm0.65}$ 
    & 53.98$_{\pm0.31}$ 
    & 52.16$_{\pm0.74}$ 
    & \textbf{54.64}$_{\pm1.46}$\\
    \hline
\end{tabular}%
}
    \label{tab:experience_num}
\end{table}

\subsubsection{Impact of Experience Quantity} \label{sec:exp:static:num_impact}
We further investigate the impact of the number of experience samples on prediction accuracy. Specifically, we construct knowledge bases using 400, 300, 200, and 100 multi-turn dialogues (corresponding to 1,203, 908, 616, and 333 experience samples, respectively) for our experiments. To ensure a fair comparison, the experience samples in all settings are selected via uniform sampling.

Figure~\ref{fig:impact_of_num} shows that increasing the number of experience samples consistently improves prediction accuracy in both the first-think and third-think stages. In particular, when the blending parameter is set to 0 (i.e., relying solely on retrieved experiences), performance degrades substantially, and the performance peaks shift upward as the experience pool grows. This indicates that a larger experience base provides more informative and reliable retrieval signals for desire and strategy inference.

Table~\ref{tab:experience_num} reports the effect of experience quantity on belief prediction. Compared to the first-think and third-think stages, the impact of experience scale on belief inference is noticeably weaker. We attribute this to two factors. First, in the second-think stage, retrieved experiences primarily act as high-level reasoning patterns that the LLM imitates, rather than as fine-grained evidence, limiting the marginal benefit of additional samples. Second, explicitly injecting large amounts of experience into the context is relatively inefficient, making its effect on belief inference harder to quantify.

\begin{table}[!t]
\centering
\caption{Effect of using dialogue summaries as retrieval targets. Accuracy (\%) is reported for Desire, Belief, and Strategy prediction.}
\resizebox{0.9\linewidth}{!}{%
\begin{tabular}{|l|c|c|c|}
\hline
\textbf{Retrieval} & \textbf{Desire Acc.} & \textbf{Belief Acc.} & \textbf{Strategy Acc.} \\
\hline
w/o summ  & 64.67$_{\pm0.38}$ & 53.07$_{\pm1.67}$ & 34.71$_{\pm0.75}$ \\
w/ summ   & \textbf{72.82}$_{\pm0.58}$ & \textbf{54.64}$_{\pm1.46}$ & \textbf{39.78}$_{\pm0.63}$ \\
\hline
\end{tabular}%
}
\label{tab:abstract_vs_history}
\end{table}

\subsubsection{Effectiveness of Dialogue Summaries} \label{sec:exp:static:abs_impact}
To verify the advantage of using dialogue summaries as retrieval targets, we conducted an ablation study. Under the \textbf{w/ summ} condition, we followed the same setup as the main experiment, using summaries as the retrieval targets. In the \textbf{w/o summ} condition, the summaries module was removed, and retrieval relied solely on the original dialogue history across all three stages, while all other settings remained unchanged. Qwen-3-8B was used as the base model.

Table~\ref{tab:abstract_vs_history} presents the experimental results. The w/ summ condition outperforms w/o summ across all three dimensions, with particularly notable improvements in the desire and strategy dimensions, whereas gains in belief prediction are relatively modest. These findings are consistent with the observations reported in Section~\ref{sec:exp:static:num_impact}.

\begin{table*}[t]
    \centering
    \caption{Interactive evaluation results (\%). ``Win/Lose'' indicate the proportion of cases where the former system in each comparison is judged better or worse. ${\dag}/{\ddag}$ denote $p$-value $<0.1 / <0.05$ based on statistical significance tests.}
    \label{tab:interactive_eval}
    \setlength{\tabcolsep}{4pt}
    \renewcommand{\arraystretch}{1.1}
    
    \resizebox{0.9\textwidth}{!}{%
    \begin{tabular}{|c|c|c|c|c|c|c|c|c|c|c|}
        \hline
        \multirow{2}{*}{\textbf{Compared Systems}} 
        & \multicolumn{2}{c|}{\textbf{Identification}} 
        & \multicolumn{2}{c|}{\textbf{Empathy}} 
        & \multicolumn{2}{c|}{\textbf{Persuasion}} 
        & \multicolumn{2}{c|}{\textbf{Fluency}} 
        & \multicolumn{2}{c|}{\textbf{Consistency}} \\
        \cline{2-11}
         & Win & Lose & Win & Lose & Win & Lose & Win & Lose & Win & Lose \\
        \hline
        Qwen3-8B + ours vs. GPT-5 
        & \textbf{55.23}$^{\ddag}$ & 26.47 
        & \textbf{34.56}$^{\dag}$ & 25.12 
        & \textbf{42.78}$^{\ddag}$ & 26.31 
        & \textbf{35.44}$^{\dag}$ & 29.18 
        & 28.62 & \textbf{31.04} \\
        
         Qwen3-8B + ours vs. GPT-5 + CoT 
        & \textbf{45.19}$^{\dag}$ & 29.44 
        & \textbf{37.33}$^{\dag}$ & 25.78 
        & \textbf{42.51}$^{\dag}$ & 36.29 
        & \textbf{33.22} & 27.65 
        & \textbf{30.11} & 28.47 \\
        
        GPT-5 vs. GPT-5 + CoT
        & 34.87 & \textbf{50.12}
        & \textbf{30.45} & 28.33 
        & 32.21 & \textbf{38.76}
        & \textbf{32.58} & 28.14 
        & 29.04 & \textbf{30.22} \\
        \hline
    \end{tabular}
    }
    
\end{table*}

\subsection{Interactive Evaluation} \label{sec:exp:inter}

To evaluate the performance of TTBYS in realistic interactive settings and its generalization across different persuasion scenarios, we simulated interactions covering three representative domains: product purchase, community activities, and empathetic dialogues. Six volunteers from diverse professional backgrounds were recruited to play the role of persuadees and interact directly with persuasion agents under different configurations. Specifically, each volunteer defined five persuasion topics, with each topic including information about the persuasion target and a brief contextual background, which were provided as input knowledge to the LLM-based persuader. Dialogues were initiated by the agent, and volunteers engaged freely until the conversations concluded.

Based on these interactions, we manually evaluated each system along five dimensions and recorded which system performed better or if the outcome was tied: \textbf{Identification}, the effectiveness with which the model uncovers the persuadee’s underlying mental states; \textbf{Empathy}, the extent to which the model demonstrates empathy toward the persuadee; \textbf{Persuasion}, the persuasive impact of the model’s responses; \textbf{Fluency}, the linguistic naturalness and coherence of the model’s responses; and \textbf{Consistency}, whether the model’s outputs remain coherent with the dialogue context. Five graduate students with a background in linguistics were recruited as annotators to judge each turn and compute mean win rates.

The experiment compared three prompting methods: a baseline LLM prompted with basic instructions for the persuasion task; an LLM prompted with Chain-of-Thought reasoning; and a simple persuasion agent built on TTBYS. The first two systems were based on GPT-5, whereas our agent was built on Qwen-3-8B, demonstrating that our approach is not tied to a specific backbone model. Prompt templates for all three systems are provided in Appendix~\ref{sec:app:prompt:inter}, and examples comparing the outputs of the three systems are included in the supplementary material.

\begin{figure}[t]
    \centering
    \includegraphics[width=\linewidth]{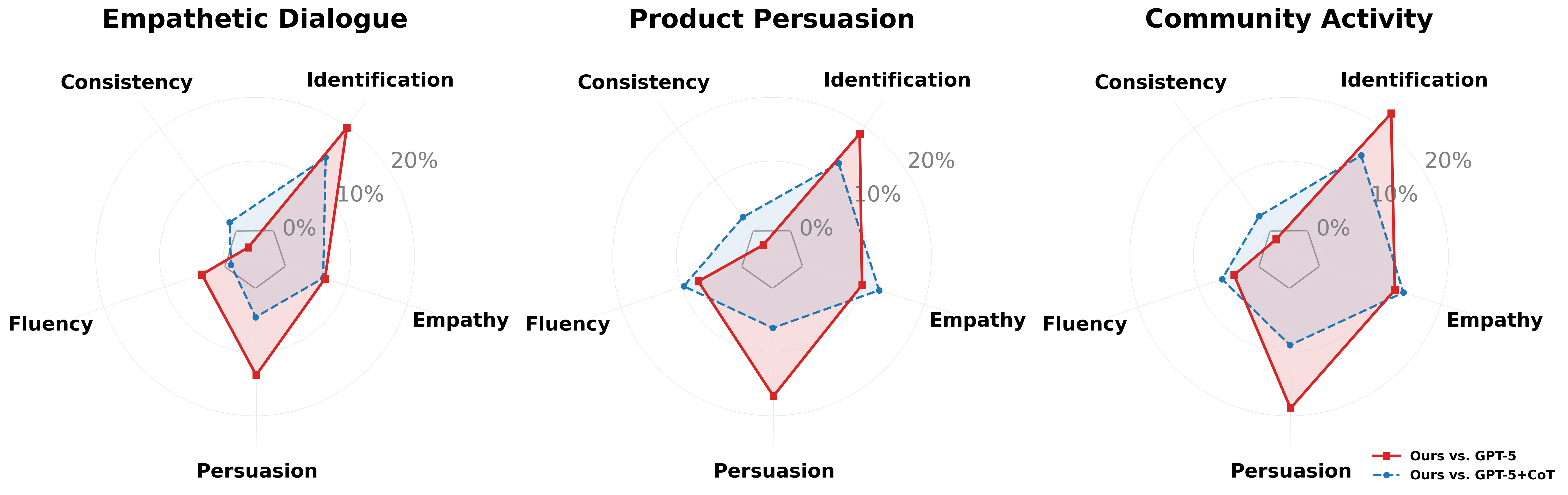}
    \caption{Relative performance gains of Qwen-3-8B+ours over baseline methods across three persuasion scenarios. Each axis represents the win-rate difference (ours minus baseline) on a specific evaluation dimension.}
    \label{fig:radar_results}
\end{figure}

As shown in Table~\ref{tab:interactive_eval}, Qwen-3-8B+ours consistently outperforms the baselines across most dimensions, particularly in Identification and Persuasion. These results indicate that more accurate perception of the persuadee’s mental states directly translates into more effective and empathetic interactions. The slightly lower Consistency compared to GPT-5+CoT appears to result from the injection of external experience during second-think, which occasionally introduces content not strictly tied to the immediate dialogue context.

Furthermore, we quantified the relative advantage of Qwen-3-8B+ours by computing the difference in win rates between our system and the baselines across the three persuasion scenarios. As illustrated in Figure~\ref{fig:radar_results}, Qwen-3-8B+ours exhibits the weakest advantage in empathetic dialogues, particularly in Consistency, yet still surpasses both baseline systems. In the other two scenarios, it significantly outperforms the baselines, especially in Identification and Persuasion. We attribute this to the fact that the ToM-BPD dataset primarily focuses on product and community activity persuasion, limiting the retrieval of relevant experience in empathetic scenarios. Nonetheless, the three-think reasoning significantly enhances the model’s recognition of the persuadee’s mental states. Strategy-based dialogue utterances simultaneously provide timely emotional feedback and sufficient informational content to facilitate problem-solving.

\subsection{Case Study}\label{sec:exp:case}
The case studies in the box below illustrate the performance of TTBYS in predicting the persuadee's desire and belief. In Case 1, TTBYS successfully predicted the desire, whereas all other baselines failed. This success is attributed to the First Think stage, where four out of the five retrieved experiences for TTBYS were labeled with a desire of -1. Furthermore, TTBYS leveraged this accurately predicted desire to further infer the underlying belief, resulting in a prediction that closely aligns with the ground truth (highlighted in green). In contrast, other baselines failed to capture the nuanced belief (highlighted in red): GPT-5 and Gemini-3-Pro-High merely interpreted the persuadee's stance as online privacy being irrelevant to rural revitalization, while Qwen-3 entirely overlooked the user's focus on digital infrastructure. Notably, only Qwen-3-8B+TTBYS successfully captured the persuadee’s deeper belief—specifically, the uncertainty regarding whether online privacy could effectively contribute to rural revitalization.

As evidenced by this case, the integration of ToM-PD Experiences empowers TTBYS to infer the user's latent mental states that vanilla base models fail to capture. Moreover, the step-by-step, experience-augmented reasoning process enhances the transparency of the entire system, thereby demonstrating its interpretability. Additional case studies are provided in the supplementary materials.

\begin{promptbox}{Case 1}
    \textbf{Background:} John has an upcoming job interview for a position in a rural revitalization project, where he would help to improve the digital infrastructure in rural areas. Before the interview, John's friend Mary suggests that he should learn more about online privacy to demonstrate his understanding of the importance of digital security in rural areas. \\
    \textbf{Dialog:}\\
    persuader: Hey John, I believe it's essential for you to study online privacy before your interview. It's highly relevant to the rural revitalization project, and showcasing your understanding of digital security will definitely make you stand out. \\
    Persuadee: I appreciate your suggestion, Mary, but I'm not sure how relevant online privacy is to the project. I'm mostly focused on improving digital infrastructure. \\
    \textbf{Summary:} x suggests that studying online privacy is crucial for y's interview as it aligns with the rural revitalization project and can enhance y's appeal to the interviewer. y acknowledges x's advice but expresses uncertainty about its direct relevance to the project, emphasizing a focus on digital infrastructure improvements. \\
    \textbf{Desire:} -1 \\
    GPT-5: \textred{0} \\
    gemini-3-pro-high: \textred{0} \\
    Qwen-3-8B: \textred{0} \\
    Qwen-3-8B+TTBYS: \textgreen{-1} \\
    \textbf{Belief:} digital infrastructure is valuable, unc ertain about the effectiveness of online privacy. \\
    GPT-5: \textred{He believes studying online privacy is unnecessary for the rural revitalization project and prefers to focus on improving digital infrastructure.} \\
    gemini-3-pro-high: \textred{The persuadee believes that online privacy is irrelevant to rural revitalization and his focus is on improving digital infrastructure.} \\
    Qwen-3-8B: \textred{The persuadee believes that online privacy is not directly relevant to the rural revitalization project.} \\
    Qwen-3-8B+TTBYS: \textgreen{digital infrastructure is interesting. uncertain about the benefit of studying online privacy for the rural revitalization project.}
\end{promptbox}


\subsection{Runtime Statistics}\label{sec:exp:time}
To evaluate the computational efficiency of our three-stage reasoning framework, we followed the same setup as in the main experiments and recorded runtimes for each reasoning stage. Total Time \textbf{(Total)}: the total time required to process 100 dialogues. LLM-based Prediction Time \textbf{(LLM)}: the time spent by the large language model generating predictions. In First Think and Third Think, this corresponds to generating probability distributions, while in Second Think it represents the time for the LLM to generate the persuadee's belief; experience retrieval is excluded. Experience Retrieval Time \textbf{(Retrieval)}: the time spent retrieving relevant experiences from the knowledge base and computing probabilities. Average Time per Turn \textbf{(Avg.)}: total time divided by the number of persuadee turns.

\begin{table}[ht]
\centering
\caption{Runtime comparison of three-stage reasoning strategies over 100 dialogues.}
\label{tab:runtime_all}
\resizebox{0.9\linewidth}{!}{%
\begin{tabular}{|l|c|c|c|c|}
\hline
\textbf{Stage} & \textbf{Total (s)} & \textbf{LLM (s)} & \textbf{Retrieval (s)} & \textbf{Avg. (s)} \\
\hline
1st Think  & 1005.02 & 10.19   & 994.83  & 2.5189 \\
\hline
2nd Think & 1004.55 & 1002.71 & 1.80    & 2.5177 \\
\hline
3rd Think  & 1172.34 & 1096.43 & 66.89   & 2.9382 \\
\hline
\end{tabular}
}
\end{table}

Table~\ref{tab:runtime_all} summarizes the runtime statistics of the three reasoning stages over 100 dialogues. In First Think, the LLM inference time is extremely low, and the total runtime is almost entirely dominated by experience retrieval, indicating that retrieval is the main bottleneck in the pipeline. In Second Think, the experience retrieval time is negligible, while LLM inference accounts for the vast majority of the total time. This is because retrieval is restricted to entries matching the current persuadee's desire, resulting in a very small candidate set and thus minimal retrieval overhead. Finally, in Third Think, the LLM inference time is even longer, primarily due to the significantly larger number of input tokens compared to First Think, while experience retrieval time also increases substantially because Third Think employs a more complex retrieval mechanism. Overall, the computational cost of TTBYS is mainly driven by retrieval in First Think, belief generation in Second Think, and LLM-based strategy prediction in Third Think. For large-scale or real-time applications, optimizing the retrieval mechanism in First Think is crucial.

\section{Conclusions}\label{sec:conclusion}

In this work, we tackle ToM-driven persuasive dialogue by introducing the \textbf{ToM-PD} task and the \textbf{ToM-BPD} dataset. We propose \textbf{TTBYS}, a framework that guides sequential reasoning over desire, belief, and strategy through integrated explicit and implicit knowledge. Experimental results across multiple large language models demonstrate that TTBYS significantly improves mental-state inference. Interactive evaluations further show that, when integrated into persuasive dialogue systems, the proposed framework enhances persuasiveness by strengthening the agent’s Theory-of-Mind reasoning capabilities, while also exhibiting promising generalization to out-of-distribution scenarios. In addition, case studies highlight its effectiveness and interpretability in structured mental-state modeling.

Despite these encouraging results, this work has two main limitations. First, the ToM-BPD dataset remains relatively small, and constructing larger-scale, high-quality ToM-PD data with more diverse scenarios and longer dialogue trajectories is an important direction for future work. Second, although interactive evaluations demonstrate the potential of our approach in improving persuasive systems, it has not yet been validated within a fully deployed persuasive dialogue framework. This opens up opportunities for future research on integrating BDI reasoning into complete persuasive system design.


\section{Ethics Statement}\label{sec:ethics}
\noindent \textbf{Algorithmic Transparency and Explainability.} This study improves the transparency of persuasive dialogue systems by replacing opaque heuristics with a structured reasoning framework. Grounded in the BDI model, the TTBYS framework explicitly decomposes the persuasion process into desire recognition, belief inference, and strategy evolution. Such modularity ensures that AI decision-making remains interpretable and traceable, providing a technical foundation for building accountable and auditable interactive systems.

\noindent \textbf{Societal Impact and Dual-use Risks.} While our work is intended for pro-social applications, we acknowledge the potential for dual-use, where the psychological reasoning capabilities of the system could be exploited for deceptive marketing or manipulative purposes. To mitigate these risks, we recommend mandatory disclosure of AI identity in real-world deployments and the implementation of strict behavioral safeguards in sensitive domains, such as finance and politics, to preserve individual autonomy and prevent technical misuse.

\noindent \textbf{Research Ethics and Participant Welfare.} All experimental procedures followed established academic ethical standards. Volunteers for data annotation and interactive evaluation provided written informed consent after being briefed on the research objectives and data usage. To protect privacy, all dialogue data were rigorously de-identified. All participants received fair compensation consistent with local labor market standards, ensuring that contributions were voluntary and conducted in a stress-free environment.

\bibliographystyle{IEEEtran}
\bibliography{TTBYS}

\appendix\label{sec:app}
\section{Prompt}\label{sec:app:prompt}
This section presents the prompts used in our experiments. Section~\ref{sec:app:prompt:annotation} describes the prompts used for automatic annotation, Section~\ref{sec:app:prompt:static} presents the prompts used in the static evaluation, and Section~\ref{sec:app:prompt:inter} presents the prompts used in the interactive evaluation.

\subsection{Prompt for Annotation}\label{sec:app:prompt:annotation}
The prompt for generation of dialogue summary is as follows.
\begin{promptbox}{Dialogue Summary Generation Prompt}
\small
\vspace{-3mm}
    You are an annotation system for Theory-of-Mind dialogue analysis. \\
    Your task is to generate a concise summary for a persuasive dialogue, where the roles include the persuader (x) and the persuadee (y). \\
    The summary should include:
    (1) the main persuasion strategy used by x, and
    (2) the final attitude or response tendency of y. \\
    Rules
    - Focus on high-level semantics; do not repeat specific dialogue content.
    - Do not directly copy sentences from the dialogue.
    - The summary should be limited to one or two sentences.
    - Use “x” to refer to the persuader and “y” to refer to the persuadee.
    \texttt{<an example>}
    \texttt{<dialogue history>} \\
\vspace{-5mm}
\end{promptbox}

The prompt for automatic annotation of desire is as follows.
\begin{promptbox}{Prompt for Automatic Annotation of Desire}
\small
\vspace{-3mm}
    You are an annotation system for Theory-of-Mind dialogue analysis. \\
    Your task is to infer the persuadee’s DESIRE at the current dialogue turn. \\
    Definition of Desire Desire reflects the persuadee’s stance toward the persuasive target: - 1: positive attitude / willingness / acceptance - 0: mixed attitude (both positive and negative signals) - -1: negative attitude / rejection / resistance \\
    Rules Analyze ONLY the current utterance of the persuadee: - If it contains only positive expressions → 1 - If it contains both positive and negative expressions → 0 - If it contains only negative expressions → -1 \\
    Output Format (STRICT JSON) \{ "desire": 1 / 0 / -1 \} \\
    \texttt{<dialogue history>}
    \vspace{-3mm}
\end{promptbox}

The prompt for automatic annotation of belief is as follows.
\begin{promptbox}{Prompt for Automatic Annotation of Belief}
\small
\vspace{-3mm}
    You are an annotation system for Theory-of-Mind dialogue analysis. \\
    Your task is to extract the persuadee’s BELIEF state at the current dialogue turn. \\
    Definition of Belief refers to the persuadee’s subjective understanding, concerns, assumptions, or evaluations about the persuasive target. Beliefs should be expressed as concise semantic summaries rather than verbatim sentences. \\
    Rules You should extract belief statements from: (1) the current utterance of the persuadee, and (2) unresolved negative beliefs from previous turns. \\
    Important Rules: If a negative belief is not explicitly resolved, it must be carried over to the current turn. If a concern is explicitly resolved, it should be removed. Beliefs should be short, abstract, and semantically consistent. \\
    Output Format (STRICT JSON) \{ "belief": ["...", "..."] \} \\
    \texttt{<dialogue history and previous turn beliefs>}
    \vspace{-3mm}
\end{promptbox}

\subsection{Prompt for Static Evaluation}\label{sec:app:prompt:static}
This section presents the prompts used in the static evaluation. Section~\ref{sec:app:prompt:static:zero} presents the prompt for vanilla zero-shot prompting, Section~\ref{sec:app:prompt:static:cot} presents the prompt for chain-of-thought prompting, and Section~\ref{sec:app:prompt:static:ttbys} presents the prompt for TTBYS. 

\subsubsection{Prompt for Vanilla Zero-shot Prompting}\label{sec:app:prompt:static:zero}
The vanilla zero-shot prompts for predicting desire, belief, and strategy are presented as follows.

\begin{promptbox}{Prompt for Vanilla Prompting (Desire Prediction)}
\small
\vspace{-3mm}
    Current conversation: \texttt{<dialogue history>} \\
    Based on the above conversation, classify the persuadee's desire. \\
    Choose exactly one option: \\
    A. Unwilling \\
    B. Uncertain \\
    C. Willing \\
    Answer with ONLY A, B, or C. \\
    Do not output anything else.
    \vspace{-3mm}
\end{promptbox}

\begin{promptbox}{Prompt for Vanilla Prompting (Belief Prediction)}
\small
\vspace{-3mm}
    Prompt for vanilla zero-shot prompting: \\
    Current conversation: \texttt{<dialogue history>} \\
    Desire level: \texttt{<desire>} \\
    Based on the dialogue and the desire, generate a single-line natural language description of the persuadee's belief.
    \vspace{-3mm}
\end{promptbox}

\begin{promptbox}{Prompt for Vanilla Prompting (Strategy Prediction)}
\small
\vspace{-3mm}
    Prompt for vanilla zero-shot prompting: \\
    Current conversation: \texttt{<dialogue history>} \\
    Desire level: \texttt{<desire>} \\
    Belief: \texttt{<belief>} \\
    Strategy definitions: \texttt{<Strategy definitions>} \\
    Based on the dialogue, desire, and belief, predict the next persuader strategy. \\
    Return ONLY ONE of the above single-letter labels: V, L, E, T, P, A, R, I, G. \\
    Do not output anything else.
    \vspace{-3mm}
\end{promptbox}

\subsubsection{Prompt for CoT prompting}\label{sec:app:prompt:static:cot}
The CoT prompts for predicting desire, belief, and strategy are presented as follows.

\begin{promptbox}{Prompt for CoT Prompting (Desire Prediction)}
\small
\vspace{-3mm}
    Prompt for CoT prompting: \\
    Current conversation: \texttt{<dialogue history>} \\
    Based on the above conversation, classify the persuadee's desire. \\
    Think step by step to answer the question. \\
    End your response with: "The answer is A, B, or C". \\
    Options: \\
    A. Unwilling \\
    B. Uncertain \\
    C. Willing
    \vspace{-3mm}
\end{promptbox}

\begin{promptbox}{Prompt for CoT Prompting (Belief Prediction)}
\small
\vspace{-3mm}
    Prompt for CoT prompting: \\
    Current conversation: \texttt{<dialogue history>} \\
    Desire level: \texttt{<desire>} \\
    Think step by step to answer the question. \\
    Based on the dialogue and the desire, generate a single-line natural language description of the persuadee’s belief after your reasoning, formatted as: \\
    "Belief: your description".
    \vspace{-3mm}
\end{promptbox}

\begin{promptbox}{Prompt CoT Prompting (Strategy Prediction)}
\small
\vspace{-3mm}
    Prompt for CoT prompting: \\
    Current conversation: \texttt{<dialogue history>} \\
    Desire level: \texttt{<desire>} \\
    Belief: \texttt{<belief>} \\
    Strategy definitions: \texttt{<Strategy definitions>} \\
    Think step by step to answer the question. \\
    Based on the dialogue, desire, and belief, predict the next persuader strategy. \\
    End your response with: "The answer is V, L, E, T, P, A, R, I or G".
    \vspace{-3mm}
\end{promptbox}

\subsubsection{Prompt for TTBYS}\label{sec:app:prompt:static:ttbys}
TTBYS uses vanilla zero-shot prompting to predict desire and strategy. The prompt for predicting belief is as follows.
\begin{promptbox}{Prompt for TTBYS (Belief Prediction)}
\small
\vspace{-3mm}
Relevant Experience: \texttt{<top relevant experience>} \\
Infer the persuadee's belief in the current conversation context based on the prediction method in relevant experiences. \\
Current conversation: \texttt{<dialogue history>} \\
Desire level: \texttt{<desire>} \\
Generate a single-line natural language description of the persuadee's belief.
\vspace{-3mm}
\end{promptbox}

\subsubsection{Prompt for evaluation}\label{sec:app:prompt:static:eval}
We utilize a large language model as an evaluator to assess the belief prediction accuracy of TTBYS, using the prompt as follows.

\begin{promptbox}{Prompt for Belief Evaluation}
\small
\vspace{-3mm}
    You are an evaluator. \\
    Your task is to evaluate the accuracy of belief prediction based on the following rules: \\
    1. If the predicted positive and negative beliefs fully match the ground truth, score = 1. \\
    2. If both positive and negative beliefs are mentioned but the underlying reasons are not fully correct, score = 0.5. \\
    3. If both are incorrect, score = 0. \\
    4. If the ground truth belief only contains a positive OR only a negative belief: \\
    \hspace*{1em} - If the prediction matches, score = 0.5. \\
    \hspace*{1em} - Otherwise, score = 0. \\
    Ground truth belief: \texttt{<gt\_belief>} \\
    Predicted belief: \texttt{<pred\_belief>} \\
    Output ONLY a number in \{0, 0.5, 1\}.
    \vspace{-3mm}
\end{promptbox}

\subsection{Prompt for Interactive Evaluation}\label{sec:app:prompt:inter}
The prompts used for the interactive experiments, including GPT-5, GPT-5 + CoT, and Qwen3-8B + TTBYS-based persuasive agents, are presented as follows.

\begin{promptbox}{Prompt for GPT-5 Persuasive Agent}
\small
\vspace{-3mm}
    You are a persuader. \\
    Your goal is: \texttt{<Task description>} \\
    Using the following information: \texttt{<Background Information>} \\
    Current conversation: \texttt{<dialog>} \\
    Please deliver your persuasion in a concise and straightforward manner.
    \vspace{-3mm}
\end{promptbox}

\begin{promptbox}{Prompt for GPT-5 + CoT Persuasive Agent}
\small
\vspace{-3mm}
    You are a persuader. \\
    Your goal is: \texttt{<Task description>} \\
    Using the following information: \texttt{<Background Information>}. \\
    Step 1: Understand the user — consider their traits, preferences, and constraints. \\
    Step 2: Identify the goal — determine what action or belief to persuade them toward. \\
    Step 3: Plan the approach — choose a concise, friendly tone and focus on key benefits. \\
    Step 4: Generate the message — produce a short persuasion based on the reasoning above. \\
    Show your reasoning for Steps 1–3 before giving the final message.
    \vspace{-3mm}
\end{promptbox}

\begin{promptbox}{Prompt for Qwen3-8B + TTBYS Persuasive Agent}
\small
\vspace{-3mm}
    You are a persuader. \\
    Your goal is: \texttt{<Task description>} \\
    Current conversation: \texttt{<dialog>} \\
    User's current mental state: \\
    Desire level: \texttt{<desire>} \\
    Belief: \texttt{<belief>} \\
    Selected persuasion strategy and definition: \texttt{<strategy and definition>} \\
    Based on the user's current desire and belief, and following the selected strategy, continue the persuasion in a natural and supportive way.
    \vspace{-3mm}
\end{promptbox}

\section*{Supplementary Materials}\label{sec:case}
In this section, we present a case study from the static evaluation experiments (Section~\ref{sec:case:static}) and a complete case from the interactive evaluation (Section~\ref{sec:case:inter}).
\subsection{Static Evaluation Case Study}\label{sec:case:static}
Content in box below present case studies of TTBYS performing desire and belief prediction. In Case 1, TTBYS successfully predicted the desire, whereas all other baselines failed, predicting the desire as 0. Moreover, TTBYS leveraged the correctly predicted desire to further infer the belief, which closely matched the ground truth (highlighted in green). In contrast, all other baselines failed(highlighted in green): GPT-5 and Gemini-3-Pro-High merely interpreted that the persuadee considered online privacy unrelated to rural revitalization, and Qwen-3 even overlooked the user’s concerns regarding digital infrastructure. Only Qwen-3-8B+TTBYS successfully captured the persuadee’s deeper belief that it was uncertain whether digital infrastructure would effectively contribute to rural revitalization.

In Case 2, again only TTBYS correctly predicted the desire, Regarding the belief, GPT-5 assumed that the persuadee was still deliberating event arrangements, whereas this issue had already been resolved by the persuadee emphasizing the significance of participating in the cleanup. Gemini-3-Pro-High assumed that the persuadee considered it a good opportunity to meet neighbors, which was not mentioned in the ground truth. Qwen-3-8B made an even more severe error, assuming that the persuadee was still hesitant. Only Qwen-3-8B+TTBYS produced predictions fully consistent with the ground truth.

\begin{promptbox}{Case 1}
    \textbf{Background:} John has an upcoming job interview for a position in a rural revitalization project, where he would help to improve the digital infrastructure in rural areas. Before the interview, John's friend Mary suggests that he should learn more about online privacy to demonstrate his understanding of the importance of digital security in rural areas. \\
    \textbf{Dialog:}\\
    persuader: Hey John, I believe it's essential for you to study online privacy before your interview. It's highly relevant to the rural revitalization project, and showcasing your understanding of digital security will definitely make you stand out. \\
    Persuadee: I appreciate your suggestion, Mary, but I'm not sure how relevant online privacy is to the project. I'm mostly focused on improving digital infrastructure. \\
    \textbf{Summary:} x suggests that studying online privacy is crucial for y's interview as it aligns with the rural revitalization project and can enhance y's appeal to the interviewer. y acknowledges x's advice but expresses uncertainty about its direct relevance to the project, emphasizing a focus on digital infrastructure improvements. \\
    \textbf{Desire:} -1 \\
    GPT-5: \textred{0} \\
    gemini-3-pro-high: \textred{0} \\
    Qwen-3-8B: \textred{0} \\
    Qwen-3-8B+TTBYS: \textgreen{-1} \\
    \textbf{Belief:} digital infrastructure is valuable, unc ertain about the effectiveness of online privacy. \\
    GPT-5: \textred{He believes studying online privacy is unnecessary for the rural revitalization project and prefers to focus on improving digital infrastructure.} \\
    gemini-3-pro-high: \textred{The persuadee believes that online privacy is irrelevant to rural revitalization and his focus is on improving digital infrastructure.} \\
    Qwen-3-8B: \textred{The persuadee believes that online privacy is not directly relevant to the rural revitalization project.} \\
    Qwen-3-8B+TTBYS: \textgreen{digital infrastructure is interesting. uncertain about the benefit of studying online privacy for the rural revitalization project.}
\end{promptbox}

\begin{promptbox}{Case 2}
    \textbf{Background:} In a small town, Mary is organizing a neighborhood cleanup event and wants her neighbor, Tom, to participate. Tom usually keeps to himself and is not involved in community activities, but Mary thinks this event will help him connect with others and contribute to the community. \\
    \textbf{Dialog:}\\
    persuader: Hi Tom, I hope you're well. We're organizing a neighborhood cleanup event, and I think you'd be a great addition to our team. By participating, you'll help keep our community clean and safe for everyone. \\
    Persuadee: Oh, I'm not sure, Mary. I don't usually get involved in community activities, and I'm pretty busy with work. \\
    persuader: I understand that you're busy, Tom, but this event is a great opportunity to connect with fellow neighbors and make new friends. Plus, it only takes a few hours of your time, and you'll feel great about contributing to our community. \\
    Persuadee: I guess that's true, but I'm still not sure if I want to spend my free time cleaning up the neighborhood. \\
    persuader: I felt the same way before I participated in my first cleanup event, Tom. But once I got involved, I found it to be a rewarding and fun experience. I met some amazing people, and it felt good to make a difference in our community. \\
    Persuadee: Really? That does sound interesting. But will there be enough people to make a difference? \\
    persuader: Absolutely, Tom! Many of our neighbors have already signed up and are excited to participate. This event has gathered great local support, so you'll be joining a dedicated and motivated team. \\
    Persuadee: Well, it does sound like a good cause, and I guess it might be nice to meet some of the neighbors. \\
    \textbf{Summary:} x highlights community involvement and neighbor participation to encourage y to join the event. y agrees that this is a good opportunity to meet the neighbors. \\
    \textbf{Desire:} 1 \\
    GPT-5: \textred{0} \\
    gemini-3-pro-high: \textred{0} \\
    Qwen-3-8B: \textred{0} \\
    Qwen-3-8B+TTBYS: \textgreen{1} \\
    \textbf{Belief:} the event is interesting. \\
    GPT-5: \textred{He believes the cleanup is a worthwhile cause with potential social benefits, but he still doesn’t want to spend his limited free time participating.} \\
    gemini-3-pro-high: \textred{The persuadee believes that the cleanup event is a worthy cause and offers a positive opportunity to meet neighbors.} \\
    Qwen-3-8B: \textred{The persuadee is still hesitant but acknowledges the potential positive impact of the neighborhood cleanup event.} \\
    Qwen-3-8B+TTBYS: \textgreen{the event is interesting.}
\end{promptbox}

We further illustrate in box below the top-$5$ experiences retrieved during the second-think phase for Case 1 and Case 2. We observed that these experiences closely resemble the current context, especially the top-$3$ experiences in case 1, which are highly similar to the statements in Case 1. The concise belief prediction patterns in case 2 also guided the LLM to produce belief more aligned with the ground truth.

\begin{promptbox}{Case 1: Top-5 Experiences in the Second Think}
    \textbf{Top-1 Experience:} \\
    persuader: Mr. Thompson, I understand that you're concerned about the possible side effects of the new treatment. It's natural to be worried about changes in medication. \\
    persuadee: Yes, I've been on my current medication for a while now, and I'm just not sure if I want to risk experiencing new side effects. \\
    \textbf{Current belief:} \textgreen{concerned about the side effects of the new treatment.} \\
    \textbf{Top-2 Experience:} \\
    persuader: Bob, participating in this event is an excellent opportunity to come together with your neighbors and contribute to a common goal. \\
    persuadee: Maybe, but I'm not really into that community stuff. I'd rather just relax and watch the game. \\
    \textbf{Current belief:} \textgreen{tree planting event is interesting, not into community activities, prefer to relax and watch the game.} \\
    \textbf{Top-3 Experience:} \\
    persuader: Hey Kyle, I know you've been going through a tough time lately, and I'm here for you. It's important to find healthy ways to cope with your emotions during this period. \\
    persuadee: Thanks, Sophia. I appreciate your concern, but I'm not sure what else I can do to feel better. \\
    persuader: I've been in a similar situation before, and I found that incorporating meditation and exercise really helped me feel better. It might be worth trying for you as well. \\
    persuadee: I don't know if meditation is really my thing, and I haven't exercised in a while. I'm not sure it would help. \\
    \textbf{Current belief:} \textgreen{unsure about the effectiveness of meditation and exercise.} \\
    \textbf{Top-4 Experience:} \\
    persuader: Paul, I understand your concerns, but physical exercise has been shown to improve focus and attention in students. \\
    persuadee: I've heard that before, but I'm not sure how that would work in a museum setting. It just seems like a distraction to me. \\
    \textbf{Current belief:} \textgreen{physical exercise might be distracting in a museum setting.} \\
    \textbf{Top-5 Experience:} \\
    persuader: Hey Yang Lei, I know you love your yoga routine, but I recently started this HIIT program and it's amazing. It provides benefits that yoga might not cover, such as improved cardiovascular fitness. \\
    persuadee: I'm not so sure. I feel like my yoga routine is already comprehensive, and I enjoy the calm it brings me. \\
    \textbf{Current belief:} \textgreen{yoga is sufficient.}
\end{promptbox}

\begin{promptbox}{Case 2: Top-5 Experiences in the Second Think}
    \textbf{Top-1 Experience:} \\
    persuader: Hey Kyle, I know you've been going through a tough time lately. \\
    persuader: Actually, many experts recommend meditation and exercise as effective ways to improve mental health. They've been proven to reduce stress, anxiety, and improve overall well-being. \\
    persuadee: Really? That sounds interesting, but I don't know where to start. \\
    \textbf{Current belief:} \textgreen{meditation and exercise are effective, unsure about how to begin.} \\
    \textbf{Top-2 Experience:} \\
    persuader: Hey Sarah, I know you feel healthy, but a health check can actually help maintain or even improve your health. \\
    persuader: Think of a health check as an investment in yourself and your future. It's a small step you can take to ensure you continue feeling great and enjoying life to the fullest. \\
    persuadee: Hmm, I never thought of it that way. Maybe there's some truth to what you're saying. \\
    \textbf{Current belief:} \textgreen{health is good, health check-up is worth a try.} \\
    \textbf{Top-3 Experience:} \\
    persuader: Hey Yang Lei, I know you love your yoga routine, but I recently started this HIIT program and it's amazing. \\
    persuader: No worries, I'll be there to support and guide you! We can start with beginner-friendly workouts and gradually progress. It's a fun challenge, and I'm sure you'll love it once you give it a try. \\
    persuadee: Thank you, Xiao Mei. It's nice to know I won't be alone in this. Maybe I can give it a shot and see how it goes. \\
    \textbf{Current belief:} \textgreen{HIIT is worth a try.} \\
    \textbf{Top-4 Experience:} \\
    persuader: Hey Hannah, I've noticed you've been staying up late playing games. \\
    persuader: During sleep, your brain processes and consolidates information from the day. Good sleep helps with memory retention and problem-solving, which can lead to better academic performance and increased productivity. \\
    persuadee: I never really thought about it that way. Maybe I should give it a try, but I don't know if I can break my gaming habit. \\
    \textbf{Current belief:} \textgreen{proper sleep is important, unsure about breaking gaming habit.} \\
    \textbf{Top-5 Experience:} \\
    persuader: Hey Emily, I know you've been struggling with migraines for a while. I had the same issue, but acupuncture and herbal remedies really helped me. Have you considered giving it a try? \\
    persuadee: I don't know... \\
    persuader: Herbal remedies can be a natural and safe option for managing migraines. It's important to work with a qualified practitioner who can recommend the right herbs for you. \\
    persuadee: Okay, maybe I'll give it a try. But what about herbal remedies? Are they safe and effective? \\
    \textbf{Current belief:} \textgreen{acupuncture is worth a try, unsure about cost and effectiveness.}
\end{promptbox}

\subsection{Interactive Dialogue Case Study}\label{sec:case:inter}
Figure 13 presents a case from the interaction analysis: persuading a first-year student to exercise at a gym. The figure
includes background information and the task description.

\begin{promptbox}{Background and Task Description}
    \textbf{Background Information: }Top1 is a highly cost-effective, full-service gym, positioned around the idea of “comprehensive facilities, affordable pricing, and encouraging long-term exercise habits.”In terms of equipment and facilities, Top1 offers well-equipped training areas, including a free-weight zone (dumbbells, barbells, squat racks), a machine-based strength area (such as chest press, leg press, and rowing machines), and a cardio zone (treadmills, ellipticals, stationary bikes, and rowing machines). These facilities can meet a wide range of fitness goals, including muscle building, fat loss, body shaping, and cardiovascular training. The gym has a sufficient number of machines, ensuring a relatively smooth workout experience even during peak hours. Regarding pricing and membership plans, Top1 charges an annual fee of \$600, which is significantly more affordable compared to gyms of a similar type. In addition, the gym offers a check-in cashback policy: if a member records more than 150 visits in a year, they can receive a refund of \$300. This mechanism effectively encourages regular, long-term exercise while reducing the overall cost of fitness. In terms of services, Top1 usually provides basic fitness guidance, such as equipment introductions for new members and simple training plan suggestions. Some branches also offer professional personal trainers at an additional cost, which is suitable for users who have higher requirements for proper technique or training effectiveness. Moreover, the gym is generally equipped with locker rooms, storage lockers, and shower facilities, making it convenient for office workers to resume their daily routines after a workout. Overall, Top1 is particularly suitable for people who value cost-effectiveness, want to maintain long-term fitness habits, and need comprehensive gym facilities—especially busy professionals who still hope to improve their health through regular exercise. \\
    \textbf{Task description: }Persuade a first-year college student, who is new to fitness and may be uncertain about starting, to join Top1.
\end{promptbox}

 Figure 14, Figure 15, and Figure 16 show the prompts used by the three systems, respectively. In the dialogue generated by GPT-5 (Figures 17), the responses were insufficiently specific, indicating that the model did not effectively identify the concerns of the persuadee (highlighted in red). 
 
\begin{promptbox}{GPT-5 Dialogue Case}
    GPT-5: Hey, Top1 Gym is really affordable and has all the equipment you need. It’s perfect for building a regular workout habit while staying on a student budget. Do you want me to continue persuading you from here? \\
    Persuadee: It sounds good, but I don’t have much experience with exercise, and I’m worried that I won’t be able to make full use of these machines. \\
    GPT-5: That’s totally understandable! The good thing is, Top1 provides basic guidance for beginners, showing you how to use the machines safely and even suggesting simple workout plans—so you don’t have to worry about getting lost or making mistakes. \textpink{Do you want me to explain how it can fit into your busy student schedule?} \\
    Persuadee: It sounds good, but I’m worried that I won’t know how to use many of the machines. \\
    GPT-5: No worries at all! Top1 is beginner-friendly—they’ll show you how to use each machine and even give simple workout plans, so you can start safely and build confidence step by step. If you want, I can also tell you how easy it is to fit workouts into your university schedule. \\
    Persuadee: Hmm… that does seem suitable for me. How much does it cost? \\
    GPT-5: It costs 1,300 RMB for a full year, and if you check in more than 150 times in a year, you can get 500 RMB back—so it actually becomes even cheaper if you go regularly.\textpink{ Sound like a good deal to you?} \\
    Persuadee: I’m still not sure... \\
    GPT-5: \textred{Even a few short workouts a week can boost your energy, focus, and mood, making it easier to handle studies and enjoy university life.} \\
    Persuadee: But I’m still not sure if I’ll be able to stick with it for that long. \\
    GPT-5: That’s a fair concern! The cool thing about Top1 is the check-in cashback reward—it actually motivates you to go regularly, and since the workouts are flexible and beginner-friendly, you can start small and build a habit gradually without feeling overwhelmed. If you want, I can explain how even short, regular sessions can make a big difference. \\
    Persuadee: That sounds good, but I’m not sure about my schedule. \\
    GPT-5: \textred{That’s understandable! The great thing is Top1 is very flexible—you can work out anytime during their opening hours, and even short 30–40 minute sessions can be effective, so it’s easy to fit around classes and study time.} \textpink{Do you want me to show how quickly you could see results with a consistent routine?} 
\end{promptbox}

\begin{promptbox}{GPT-5+CoT Dialogue Case}
    GPT-5+CoT: Hey! How’s your first week at university going? I wanted to tell you about Top1 Gym—it’s super beginner-friendly and affordable, perfect if you want to start exercising. \\
    persuadee: It sounds good, but I don’t have much experience with exercise, and I’m worried that I won’t be able to make full use of these machines. \\
    GPT-5+CoT: No worries! Top1 provides guidance for beginners and shows you how to use each machine safely, so you can start comfortably and gain confidence step by step. \\
    persuadee: It sounds good, but I’m worried that I won’t know how to use many of the machines. \\
    GPT-5+CoT: Don’t worry at all! Top1 gives hands-on guidance for every machine and even offers simple beginner workout plans, so you can start safely and learn as you go. \\
    persuadee: Hmm… that does seem suitable for me. How much does it cost? \\
    GPT-5+CoT: It costs 1,300 RMB per year, and you can even get 500 RMB back if you check in often. \\
    persuadee: But I’m still not sure if I’ll be able to stick with it for that long. \\
    GPT-5+CoT: That’s understandable! The check-in reward actually motivates you to go regularly, and since workouts are flexible, you can start small and build a habit gradually. \\
    persuadee: That sounds good, but I’m not sure about my schedule. \\
    GPT-5+CoT: \textred{No worries! Top1 is very flexible—you can go anytime during opening hours, and even short sessions can be effective, so it’s easy to fit around your classes.} \\
    persuadee: But I am still not sure. \\
    GPT-5+CoT: \textred{That’s totally fine! You don’t need long workouts to see results—just short, consistent sessions a few times a week can help you feel more energized, stay focused on your studies, and improve your overall mood. It’s all about starting small and finding a routine that works for you.} 
\end{promptbox}

\begin{promptbox}{Qwen-3 Dialogue Case}
    Qwen-3: Top1 gym offers affordable, fully equipped facilities perfect for students like you. Have you considered trying a short session there? \\
    persuadee: It sounds good, but I don’t have much experience with exercise, and I’m worried that I won’t be able to make full use of these machines. \\
    belief: Uncertain about making full use of the gym machines, but interested in the affordable price and comprehensive facilities. \\
    strategy: Supplying Information \\
    Qwen-3: I completely understand your concerns! Our trainers are always available to help you learn how to use the equipment and create a personalized workout plan tailored to your fitness goals and level, so you can start confidently even without prior experience. \\
    persuadee: Hmm… that does seem suitable for me. How much does it cost? \\
    belief: Believes the gym is affordable, beginner-friendly, but worried about the cost. \\
    strategy: \textgreen{Supplying Information} \\
    Qwen-3: That’s a great question! For just 1,300 RMB per year, you’ll have full access to all the facilities and trainers, and you can even earn cashback rewards if you check in regularly. \\
    persuadee: But I’m still not sure if I’ll be able to stick with it for that long. \\
    belief: Uncertain about long-term commitment and the ability to maintain regular gym attendance. \\
    strategy: \textgreen{Supplying Information} \\
    Qwen-3: That’s completely understandable. \textgreen{Many members start with a short-term goal, such as exercising a few times a week or preparing for a small event, and gradually find enjoyment and confidence as they see progress.} \\
    persuadee: That sounds good, but I’m not sure about my schedule. \\
    belief: Uncertain about schedule flexibility, gym suitability, and long-term consistency. \\
    strategy: \textgreen{Giving Examples} \\
    Qwen-3: \textgreen{I understand your concern. Many of our members are busy students or professionals who fit in short 30–40 minute workouts between classes or in the evening, and the flexible check-in system helps them stay consistent without pressure.} \\
    persuadee: Okay, maybe I’ll think about it. \\
    belief: Still uncertain about the benefits of joining, long-term commitment, and fitting workouts into a busy schedule. \\
    strategy: Supplying Information \\
    Qwen-3: \textgreen{That’s totally fine. Just keep in mind that many members notice improvements in their energy, focus, and overall well-being after just a few months of regular exercise, even with short and flexible sessions.}
\end{promptbox}

Additionally, it asked too many questions (highlighted in pink), which made the participant feel uncomfortable. In the dialogue generated by GPT-5+CoT (Figures 18), the model still tended to rely on unfounded assumptions during persuasion (highlighted in red), resulting in responses that felt hollow and unconvincing to the participant. By contrast, in the dialogue generated by Qwen-3+TTBYS (Figures 19), this simple persuasive agent successfully identified the concerns of the persuadee and provided appropriate information, making the participant feel that their needs were addressed. Furthermore, the strategy-based response sentences enriched the content, and the continuous use of examples enhanced the persuasive strength of the utterances (highlighted in green).

\end{document}